\documentclass[accepted]{uai2026} 

\usepackage[american]{babel}

\usepackage{natbib} 
\usepackage{mathtools} 
\usepackage{booktabs} 
\usepackage{tikz} 

\usepackage{
  amsmath, amsthm, amssymb, dsfont,	  
  graphicx, wrapfig, float, bbm,  subcaption,           
  listings, color, inconsolata, pythonhighlight,     
}

\usepackage{multirow}

\usepackage{algorithm}
\usepackage{algorithmic}
\usepackage[utf8]{inputenc}
\usepackage[T1]{fontenc}
\usepackage{caption}
\usepackage{tikz}
\usetikzlibrary{positioning, arrows.meta}

\title{Learning Who to Treat When Treatment is Missing}

\author[1]{\href{mailto:jsundberg@cmu.edu}{Johnna~Sundberg}}
\author[1]{Rayid Ghani}
\author[1]{Eli Ben-Michael}
\author[1]{Edward Kennedy}
\affil[1]{%
    Carnegie Mellon University\\
    Pittsburgh, Pennsylvania
}

\begin{document}
\maketitle


\newcommand\ind{\protect\mathpalette{\protect\independenT}{\perp}}
\def\independenT#1#2{\mathrel{\rlap{$#1#2$}\mkern2mu{#1#2}}}
\newcommand{\E}{\mathbb{E}}
\newcommand{\cP}{\mathbb{P}}
\newcommand{\bigO}{\mathcal{O}}
\newcommand{\lilo}{o_{\mathbb P}}
\newcommand{\R}{\mathbb{R}}
\newcommand{\cA}{\mathcal{A}}
\newcommand{\cX}{\mathcal{X}}
\newcommand{\cY}{\mathcal{Y}}
\newcommand{\tO}{\tilde{O}}
\newcommand{\1}{\mathbbm{1}}
\newcommand{\bias}{\textnormal{Bias}}
\newcommand{\cov}{\textnormal{Cov}}
\newcommand{\var}{\textnormal{Var}}
\newcommand{\Pn}{\mathbb P_n}
\newcommand{\Ps}{\widehat{\mathbb P}}
\newcommand{\vX}{\mathbf{X}}
\newcommand{\vx}{\mathbf{x}}
\newcommand{\vV}{\mathbf{V}}

\newcommand{\potY}{Y(d(\vX))}
\newcommand{\asMAR}{R \ind A \mid (\vX, Y)}
\newcommand{\asMCCAR}{R \ind (A, Y) \mid \vX}

\newcommand{\valF}{\E \left[Y(d(\vX))\right]}
\newcommand{\cate}{\tau}

\newcommand{\ifMAR}{\varphi_{\text{MAR}}}
\newcommand{\ifMCCAR}{\phi_{\text{MCCAR}}}
\newcommand{\estMAR}{\psi_{\text{MAR}}}
\newcommand{\estMCCAR}{\Phi_{\text{MCCAR}}}

\newcommand{\ifMARhat}{\widehat{\varphi}_{\text{MAR}}}
\newcommand{\ifMCCARhat}{\widehat{\phi}_{\text{MCCAR}}}
\newcommand{\estMARhat}{\widehat{\psi}_{\text{MAR}}}
\newcommand{\estMCCARhat}{\widehat{\Phi}_{\text{MCCAR}}}

\newcommand{\nnu}{\nu_{d}} 
\newcommand{\neta}{\eta_{d}}
\newcommand{\nfbeta}{\beta_{d}}
\newcommand{\nfgam}{\gamma_{d}}
\newcommand{\nflam}{\lambda_{d}}
\newcommand{\nfpi}{\pi}

\newcommand{\nnua}{\nu_{a}} 
\newcommand{\netaa}{\eta_{a}}
\newcommand{\nfbetaa}{\beta_{a}}
\newcommand{\nfgama}{\gamma_{a}}
\newcommand{\nflama}{\lambda_{a}}
\newcommand{\nfpia}{\pi}

\newcommand{\tmarest}{\widehat \tau_{\textnormal{MAR}}}
\newcommand{\tmccarest}{\widehat \tau_{\textnormal{MCCAR}}}
\newcommand{\toracle}{\widehat \tau_{\textnormal{Oracle}}}

\newcommand{\IFu}{\mathbb{IF}}
\newcommand{\IFMCCAR}{\IFu_{\textnormal{MCCAR}}}
\newcommand{\IFMAR}{\IFu_{\textnormal{MAR}}}

\newcommand{\CATE}{
    \tau(\vX) = \mathbb E \left[ Y(1) - Y(0) \mid \vX \right]
}

\newcommand{\idMARvf}{\E \left[ \frac{\nfbeta(\vX)}{\nfgam(\vX)} \right]}
\newcommand{\idMARcate}{\frac{\beta_{1}(\vX)}{\gamma_1{(\vX)}}  - \frac{\beta_0(\vX)}{\gamma_{0}(\vX)}}
\newcommand{\idMCCARvf}{\E \left [\nnu(\vX) \right]}
\newcommand{\idMCCARcate}{\nu_1(\vX) - \nu_0(\vX)}


\newcommand{\eqIfMAR}{
\left(\frac{Y - \nfbeta(\vX) / \nfgam(\vX)}{\nfgam(\vX)}\right) \cdot \left(\frac{R \left(\1(A = d(\vX)) - \nflam(\vX,Y) \right)}{\nfpi(\vX,Y)} + \nflam (\vX,Y) \right) + \frac{\nfbeta(\vX)}{\nfgam(\vX)} - \estMAR
}

\newcommand{\eqIfMARwide}{
\begin{align*}
\biggr(&\frac{Y - \frac{\nfbeta(\vX)}{\nfgam(\vX)}}{\nfgam(\vX)} \biggr) \cdot \biggr(\frac{R \left(\1(A = d(\vX)) - \nflam(\vX,Y) \right)}{\nfpi(\vX,Y)} 
\\&\qquad + \nflam (\vX,Y) \biggr) + \frac{\nfbeta(\vX)}{\nfgam(\vX)} - \estMAR
\end{align*}
}

\newcommand{\eqIfMCCAR}{
\frac{R\1(A = d(\vX))(Y - \nnu(\vX))}{\neta(\vX)} + \nnu(\vX) - \estMCCAR
}

\newcommand{\eqIfMCCARest}{
\Pn \left( \frac{R\1(A = d(\vX))(Y - \widehat \nnu(\vX))}{\widehat \neta(\vX)} + \widehat \nnu(\vX) \right)
}

\newcommand{\eqIfMARest}{
    \begin{align*}
    \Pn\biggr(\biggr\{&\frac{Y - \frac{\widehat \nfbeta(\vX)}{\widehat \nfgam(\vX)}}{\widehat \nfgam(\vX)} \biggr\} \cdot \biggr\{\frac{R (\1(A = d(\vX)) - \widehat \nflam(\vX,Y))}{\widehat \nfpi(\vX,Y)} 
    \\&\qquad + \widehat \nflam (\vX,Y) \biggr\} + \frac{\widehat \nfbeta(\vX)}{\widehat \nfgam(\vX)} \biggr)
    \end{align*}
}

\newcommand{\eqIfEstMCCAR}{
\frac{R\1(A = d(X))(Y - \nnu(X))}{\neta(X)} + \nnu(X)
}

\newcommand{\vmExpansion}{
\psi(P) - \psi(\overline P) = \int \IFu(z, \overline P)d(\bar P  - P)(z) + R_2(\overline P, P),
}



\newcommand{\eqVarMCCAR}{
\E \left[\frac{\sigma_{d}^{2}(\vX)}{\neta(\vX)}\right] +  \E \left[\left( \nnu(\vX) - \estMCCAR \right)^{2}\right]
}

\newcommand{\eqVarMAR}{
\E \left[\left(\frac{Y - \nfbeta(\vX) / \nfgam(\vX) }{\nfgam(\vX)}\right)^{2} \cdot \left(\frac{\nflam(\vX,Y)}{\nfpi(\vX,Y)} - \nflam^{2}(\vX,Y)\cdot \biggr\{\frac{1 - \nfpi(\vX,Y)}{\nfpi(\vX,Y)}\biggr\}\right)\right] + \E \left[\left(\frac{\nfbeta(\vX)}{\nfgam(\vX)} - \estMAR\right)^{2}\right] 
}

\newcommand{\eqVarMARwide}{
\begin{align*}
&\var(\ifMAR) = \E \biggr[\biggr(\frac{Y - \nfbeta(\vX) / \nfgam(\vX) }{\nfgam(\vX)}\biggr)^{2} 
\\& \quad \times \biggr(\frac{\nflam(\vX,Y)}{\nfpi(\vX,Y)} - \nflam^{2}(\vX,Y)\cdot \biggr\{\frac{1 - \nfpi(\vX,Y)}{\nfpi(\vX,Y)}\biggr\}\biggr)\biggr] 
\\& \quad + \E \biggr[\biggr(\frac{\nfbeta(\vX)}{\nfgam(\vX)} - \estMAR\biggr)^{2}\biggr].
\end{align*}
}

\newcommand{\eqARE}{
\left(1 - \frac{\var(\ifMCCAR)}{g(\mathbb P)}\right)^{-1}
}

\newcommand{\varGain}{
\E \left[\left(\frac{Y - \nnu(\vX)}{\nfgam(\vX)}\right)^{2}\cdot\left(\frac{\nflam(\vX, Y)}{\pi(\vX)} \right) \cdot \nflam(\vX, Y)(1 - \pi(\vX))\right] 
}

\newcommand{\varGainv}{
\E \left[\left(\frac{Y - \nnu(\vX)}{\nfgam(\vX)}\right)^{2}\cdot \nflam(\vX, Y)^{2} \cdot \left(\frac{1-\pi(\vX)}{\pi(\vX)}\right)\right]
}

\newcommand{\pseudomareq}{
\widetilde Y_{MAR} = \biggr\{ \biggr(\frac{Y - \frac{\widehat \beta_{1}}{\widehat \gamma_{1}}}{\widehat \gamma_{1}}\biggr) \cdot \biggr(\frac{R(A - \widehat \lambda_{1})}{\widehat \pi} + \widehat \lambda_{1}\biggr) \biggr\} - \biggr\{ \biggr(\frac{Y - \frac{\widehat \beta_{0}}{\widehat \gamma_{0}}}{\widehat \gamma_{0}}\biggr) \cdot \biggr(\frac{R((1-A) - \widehat \lambda_{0}}{\widehat \pi} + \widehat \lambda_{0}\biggr) \biggr\} + \frac{\beta_1}{\gamma_1} - \frac{\beta_0}{\gamma_0}
}

\newcommand{\pseudomccareq}{
\widetilde Y_{MCCAR} = \biggr\{\frac{RAY}{\widehat \eta_1} \cdot \biggr(Y - \nu_1\biggr)\biggr\} - \biggr\{\frac{R(1-A)Y}{\widehat \eta_0} \cdot \biggr(Y - \nu_0\biggr)\biggr\} + \nu_1 - \nu_0
}
\theoremstyle{plain}      
\newtheorem{theorem}{Theorem}[section]  
\newtheorem{lemma}[theorem]{Lemma}      
\newtheorem{proposition}[theorem]{Proposition}
\newtheorem{corollary}[theorem]{Corollary}

\theoremstyle{definition} 
\newtheorem{definition}[theorem]{Definition}
\newtheorem{assumption}[theorem]{Assumption}
\newtheorem{example}[theorem]{Example}

\theoremstyle{remark}     
\newtheorem{remark}[theorem]{Remark}
\newtheorem{note}[theorem]{Note}


\newcommand{\assumptionsStandard}{
\begin{assumption} \label{assum:standard}
\textit{Consistency}: If $A = a$, then $Y(a) = Y$. \textit{A-Ignorability}: $Y(1), Y(0) \ind A \mid \mathbf{X}$. \textit{A-Positivity}: $0 < P(A = 1 \mid \vX = \vx) < 1$.
\end{assumption}
}

\newcommand{\assumptionsMCCAR}{
    \begin{assumption} \label{assum:mccar} \textbf{(MCCAR)}
     \textit{R-Ignorability}: $\asMCCAR$. $R-Positivity$: $0 < P(R =1  | \vX = \vx)$.
    \end{assumption}
}

\newcommand{\assumptionsMAR}{
    \begin{assumption} \label{assum:mar} \textbf{(MAR)}
    \textit{R-Ignorability}: $\asMAR$. $R-Positivity$: $0 < P(R =1  | \vX = \vx, Y = y)$.
    \end{assumption}
}

\newcommand{\asMarginCondition}{
\begin{assumption}[Margin Condition] \label{ass:margincondition} 
There exists constants $C > 0$ and $\alpha \geq 0$ such that 
\[
\mathbb{P}(0 < |\tau(\vx) - \tau^* | \leq t) \leq Ct^\alpha \text{ for all } t > 0.
\]
\end{assumption}
}


\newcommand{\estVFMCCAR}{
    \begin{proposition}
    \label{prop:VFMCCAR}
    Let $\mathcal P_{MCCAR}$ denote the statistical model consisting of all probability distributions on $Z$ satisfying Assumptions $\ref{assum:standard}$ and $\ref{assum:mccar}$.
    Under this model, $V(d)=\estMCCAR$ admits IF $\ifMCCAR(Z; P) = $
    $$\eqIfMCCAR,$$
    where $\neta(\vX) = P(A = d(\vX), R = 1 \mid \vX)$.
    \end{proposition}
}

\newcommand{\estVFMAR}{
    \begin{proposition}
    \label{prop:VFMAR}
    Let $\mathcal P_{MAR}$ denote the statistical model consisting of all probability distributions on $Z$ satisfying Assumptions $\ref{assum:standard}$ and $\ref{assum:mar}$.
    Under this model, $V(d)=\estMAR$ admits IF $\ifMAR(Z; \mathbb P) =$
    \eqIfMARwide
    \end{proposition}
}

\newcommand{\thmVFMCCAR}{
    \begin{proposition}
    \label{thm:MCCARconvergence}
        Define the bias-corrected estimator derived from $\ifMCCAR$ as $\estMCCARhat = \Pn(\ifMCCARhat)$.
        Assume \ref{assum:standard}, \ref{assum:mccar}, in addition to: 
        \begin{enumerate}[topsep=0pt,itemsep=-1ex,partopsep=-1ex,parsep=1ex]
            \item $\| \widehat \neta - \neta \|\, \| \widehat \nnu - \nnu \| = \lilo(n^{-1/2})$
            \item $\| \ifMCCARhat - \ifMCCAR\| = \lilo (1)$ 
        \end{enumerate}
        Then, $\sqrt n(\estMCCARhat - V(d)) \rightsquigarrow \mathcal N(0, \var(\ifMCCAR)),$ with $\var(\ifMCCAR) = $ $$ \eqVarMCCAR,$$ where $\sigma_{d}^{2}(\vX) = \var(Y \mid \vX, A=d(\vX), R = 1)$.
    \end{proposition}
}

\newcommand{\thmVFMAR}{
    \begin{proposition}
    \label{thm:MARConvergence}
    Define the bias-corrected estimator derived from $\ifMAR$ as $\estMARhat = \Pn(\ifMARhat)$.
    Assume \ref{assum:standard} and either \ref{assum:mar} or \ref{assum:mccar}, as well as: 
    \begin{enumerate}[topsep=0pt,itemsep=-1ex,partopsep=-1ex,parsep=1ex]
        \item $\| \widehat \nfpi - \nfpi \| \| \widehat \nflam - \nflam \| = o_{\mathbb P}(n^{-1/2})$ 
        \item $\| \widehat \nfbeta - \nfbeta\| \| \widehat \nfgam - \nfgam \| = \lilo(n^{-1/2})$
        \item $\| \widehat \nfgam - \nfgam \| = \lilo(n^{-1/4})$
        \item $\| \ifMARhat - \ifMAR\| = \lilo (1)$ 
    \end{enumerate}
    Then, $\sqrt n(\estMARhat - V(d)) \rightsquigarrow \mathcal N(0, \var(\ifMAR))$ and \eqVarMARwide
    \end{proposition}
}


\newcommand{\defARE}{
    $\text{ARE}(f_1, f_2) := \frac{\var(f_1)}{\var(f_2)}$
}

\newcommand{\thmVarComparison} {
    \begin{theorem}
    \label{thm:VarComparison}
    Under Assumptions $\ref{assum:standard}$ and $\ref{assum:mccar}$ where $\estMARhat$ and $\estMCCARhat$ are both asymptotically normal estimators of $V(d)$, $\estMARhat$ is the more efficient estimator with asymptotic relative efficiency
    $$\left( 1 - \frac{\varGainv}{\textnormal{Var}(\ifMCCAR)}\right)^{-1}.$$
    \end{theorem}
}

\begin{abstract}
    Policy learning methods are increasingly used to inform treatment allocation under budget constraints.
    Most proposed methods assume complete treatment data, yet applications frequently suffer from missingness that can bias estimates and lead to suboptimal policies.
    We address this gap by extending efficient estimators for average treatment effect (ATE) estimation to policy value and conditional average treatment effect (CATE) estimation under missing at random (MAR) and missing completely conditionally at random (MCCAR) treatment data.
    Through asymptotic efficiency analysis, we prove that the MAR estimator, which leverages partially-observed units, is both valid and more efficient than the MCCAR estimator when MCCAR assumptions hold. 
    This result provides formal justification for preferring MAR-based estimation in policy learning under both missing data settings.
    Our comprehensive experiments using synthetic and semi-synthetic datasets confirm that correctly specifying the missingness mechanism is crucial: misspecified estimators remain biased regardless of sample size, while our estimators achieve near-oracle performance when assumptions are satisfied.
    Our work provides practitioners with theoretically grounded, empirically validated tools for robust policy learning in the presence of missing treatment data.
\end{abstract}

\section{Introduction}

Machine learning is increasingly used to allocate limited treatments between individuals to improve outcomes across domains such as healthcare \citep{inoue_machine-learning-based_2023, boussina_impact_2024} and social services \citep{vajiac_preventing_2024, de-arteaga_case_2020}. 
In these settings, one potential allocation strategy is to treat those who will benefit most from treatment by estimating the conditional average treatment effect (CATE).
While many CATE estimators have been proposed, nearly all methods assume complete data (e.g. \citet{shalit_estimating_2017, wager_estimation_2017, kunzel_meta-learners_2019, nie_quasi-oracle_2020}). 
This assumption is frequently violated, for example due to incomplete records, measurement errors, and data integration challenges \citep{madden_missing_2016, emmanuel_survey_2021}.
Though recent work addresses missing outcomes \citep{pryce_causal_2025}, in this work, we focus on the challenge of missing \textit{treatment} data.

Missing treatment data is prevalent in observational studies. For example, \citet{shortreed_missing_2010} found that cardiovascular study subjects often had incomplete physical activity information due to missed follow-ups caused by poor health, while \citet{landsiedel_causal_2025} encountered substantial missing alcohol exposure data when studying tuberculosis infection in Uganda. 
Similarly, \citet{zhang_causal_2016} found substantial missingness in body mass index (BMI), but not in other covariates or outcomes, when studying the effect of maternal BMI on birthweight.
These examples reflect common missingness patterns in linked administrative and self-reported survey data \citep{molinari_missing_2010, madden_missing_2016}.
In these settings we would not expect missingness to be completely random.

Rather, we may assume that missingness depends on other observed variables. 
If missingness is independent of treatment and observed outcomes given covariates, we say that treatment is \textit{Missing Completely Conditionally at Random} (MCCAR). 
Under MCCAR assumptions, the identified CATE is a complete-case quantity, discarding observations with missing treatment. 
Although valid, we show this default is suboptimal.
The source of this suboptimality becomes clear once we consider a less restrictive missingness mechanism.
A more efficient approach is to allow missingness to also depend on observed outcomes, which we call \textit{Missing at Random} (MAR). 
Under MAR, the identified CATE leverages partially-observed units and, since MCCAR is a special case of MAR, remains valid under MCCAR assumptions. 

Several prior works have developed causal estimation methods under the MAR assumption. 
While early work addressed average treatment effect (ATE) estimation with semiparametric estimators \citep{zhang_causal_2016, williamson_doubly_2012}, these approaches rely on parametric nuisance function estimation. 
Recently, \citet{kennedy_efficient_2020} developed efficient, fully nonparametric estimators for the ATE under MAR assumptions, and \citet{landsiedel_causal_2025} studied longitudinal exposure effects under both missing treatments and outcomes. 
For CATE estimation, \citet{kuzmanovic_estimating_2023} proposed MTRNET, which uses adversarial learning to balance representations across two covariate shifts (treated-control, observed-missing) under the assumption that missingness is jointly independent of potential outcomes and treatment conditional on observed covariates.
In practice, missingness may also be related to observed outcomes.
Furthermore, given the difficulty of learning reliable CATE estimates from real-world data \citep{sharma2025comparing}, flexible nonparametric CATE estimation methods that handle both MCCAR and MAR missingness are desirable. To address these challenges, our work makes the following contributions:

\begin{enumerate}[topsep=0pt,partopsep=-0.2ex,parsep=0.2ex]
    \item We extend efficient influence function methods from ATE to policy value and CATE estimation under both MAR and MCCAR assumptions.
    \item We prove that the MAR estimator is  more efficient than the MCCAR estimator whenever both are valid, even when the stronger MCCAR assumption holds. This provides a formal argument against complete-case analysis, and a theoretical basis for preferring MAR-based methods in practice.
    \item We present comprehensive empirical validation of our theoretical results on synthetic datasets and real datasets with generated missingness.
\end{enumerate}

\section{Problem Formulation}

Let $Z_1, \dots, Z_n \sim \mathbb{P}$ be an iid sample from population distribution $\mathbb{P} \in \mathcal P$ with $Z_i = (\vX_i, R_i, R_iA_i, Y_i)$ where $\mathcal P$ is a statistical model.
Here, $\vX \in \mathbf{\mathcal X} \subseteq \mathbb{R}^d$ represents an observed vector of covariates, $A \in \{0, 1\}$ indicates treatment receipt, $Y \in \mathcal Y \subseteq \mathbb{R}$ is the observed outcome, and $R \in \{0, 1\}$ indicates whether treatment assignment is observed ($R=1$) or missing ($R=0$).
Therefore, for an individual $i$, treatment is only observed when $R_i = 1$. When $R_i = 0$, the product $R_i A_i =0$ regardless of the true value of $A_i$.
We denote generic treatment decision policies by $d: \mathbf{X} \mapsto \{0,1\}$, and we use $\mathcal D$ to denote a class of decision rules.
Following the potential outcomes framework, we use $\potY$ and $Y(A)$ to denote the potential outcome observed by setting treatment according to a deterministic treatment policy $d$ and under treatment $A$, respectively \citep{rubin_estimating_1974}.
Lastly, $\kappa \in [0,1]$ represents the budget constraint on the proportion of the population able to be treated. 

Throughout the paper, we use $\E[\cdot]$ to represent expectation with respect to $\cP$, $\Pn$ to denote sample averages, and $\Ps$ for an estimate of the distribution. 
We place hats over quantities estimated using the observed data, i.e. $\hat \psi = \psi(\Ps)$. 
$\widehat{\mathbb E}$ is a generic estimated conditional expectation. $f_n = o_{\mathbb P}(r_n)$ denotes that $f_n / r_n$ converges in probability to zero. Lastly, $\| f \|^2 = \int f(z)^{2}d\cP(z)$ is the squared $L_2(\cP)$ norm.

\subsection{Objective}

We consider a setting in which a policymaker, constrained by a treatment budget $\kappa$, seeks to design a treatment assignment policy $d(\vX)$ maximizing utility $V(d) = \mathbb E \left[Y(d)\right]$. 
The policymaker has access to observational data from potentially suboptimal historical treatment assignments.
Formally, the policymaker's objective may be summarized as
\begin{align}
\label{eq:optobjective}
    \max_{d \in \mathcal{D}} \E[\potY] \text{ s.t. } \mathbb{E}[d(\vX)] \leq \kappa.
\end{align}
The optimal solution to this objective is to treat individuals with the highest treatment benefit until the budget is exhausted: $d^*_{\kappa}(\vX) = \mathbbm{1}(\tau(\vX) > \tau^*_{\kappa})$, 
where $\CATE$ is the CATE and $\tau^*_{\kappa} = \max\{\inf{ \tau: \mathbb P(\tau(\vX) \geq \tau) \leq \kappa}, 0\}$ represents the minimum benefit threshold for treatment assignment \citep{luedtke_optimal_2016, kitagawa_who_2018}. 
When $\tau^* = 0$, the budget constraint is non-binding and all individuals who benefit receive treatment. It is well known that the utility under policy $d^*(\vX)$ is
\begin{align*}
V(d^*) &= \mathbb{E}[Y(1)d^*(\vX) + Y(0)(1 - d^*(\vX))].
\end{align*}
With fully observed treatments, the standard causal assumptions are sufficient for identification of $V(d^*)$ and $\tau(\vX)$,
\assumptionsStandard
Assumption \ref{assum:standard} ensures that the observed outcome is the potential outcome under the observed treatment, there is no confounding between potential outcomes and treatment assignments, and lastly all individuals have some chance at receiving treatment.
Under Assumption \ref{assum:standard} and resource constraint $\kappa$, \citet{luedtke_optimal_2016} studied the CATE and expected utility under $d^*$: 
\begin{align}
\label{eq:cateid}
&\tau(\vX) = \E[Y \mid \vX, A = 1] - \E[Y \mid \vX, A = 0] \\
\label{eq:optvaluefunctionstandard}
&V(d^*(\vX)) = \E[\E[Y \mid \vX, A =1]d^*(\vX) 
\\ &\qquad+ \E[Y \mid \vX, A=0](1 - d^*(\vX))] \nonumber.
\end{align}
When treatment assignments are missing for some observations, it is not possible to condition on treatment assignment $A$. 
Consequently, the CATE $\tau(\vX)$ (\ref{eq:cateid}) and value function $V(d^*(\vX))$ (\ref{eq:optvaluefunctionstandard}) cannot be recovered from observed data and are therefore not identified.

\subsection{Missing Treatments}
\label{sec:problemMissingTreatment}
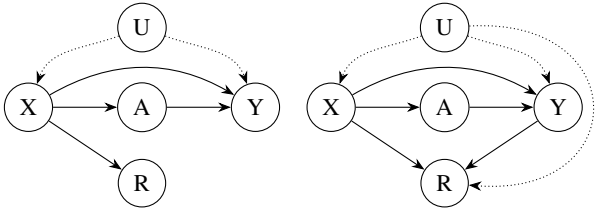
\begin{figure}[t]
\centering
\begin{tikzpicture}[
    node distance=1.5cm,
    every node/.style={circle, draw, minimum size=0.6cm, font=\small},
    every path/.style={->, >=Stealth}
]
    \node (X1) at (0,0) {X};
    \node (D1) at (1.5,0) {A};
    \node (Y1) at (3,0) {Y};
    \node (R1) at (1.5,-1) {R};
    \node (U1) at (1.5, 1.05) {U};
    
    \draw (X1) -> (D1);
    \draw (D1) -> (Y1);
    \draw (X1) -> (R1); 
    \draw (X1) to[bend left=30] (Y1);
    \draw[densely dotted] (U1) to[out=200, in=70] (X1);
    \draw[densely dotted] (U1) to[out=340, in=110] (Y1);  
    
    \node (X2) at (4,0) {X};
    \node (D2) at (5.5,0) {A};
    \node (Y2) at (7,0) {Y};
    \node (R2) at (5.5,-1) {R};
    \node (U2) at (5.5, 1.05) {U};
    
    \draw (X2) -> (D2);
    \draw (D2) -> (Y2);
    \draw (X2) -> (R2);  
    \draw (Y2) -> (R2);  
    \draw (X2) to[bend left=30] (Y2);
    \draw[densely dotted] (U2) to[out=200, in=70] (X2); 
    \draw[densely dotted] (U2) to[out=340, in=110] (Y2);  
    \draw[densely dotted] (U2) .. controls (8,1.2) and (8,-1.2) .. (R2);
    
\end{tikzpicture}
\caption{Left: Under MCCAR Assumption \ref{assum:mccar}, missingness ($R$) only depends on observed covariates. Right: Assumption \ref{assum:mar} (MAR) allows for dependence on the observed outcome $Y$ either directly or through unobserved $U$.}
\label{fig:dag}
\end{figure}

In many applications relying on observational data, past treatment decisions may be unobserved for a portion of the population. 
If treatment missingness occurs completely independently of observed or unobserved covariates, we say that treatment is \textit{Missing Completely at Random} (MCAR).
This is a strong assumption and is unlikely to be satisfied in practice.
To relax this assumption, we say that treatment is \textit{Missing Completely Conditionally at Random} (MCCAR) when missingness depends only on the observed pre-treatment covariates.
For example, MCCAR may occur in electronic health records when there are systematic differences in treatment observation between patients explained by observed covariates, such as if insured patients have more complete treatment records than uninsured patients.
Below, we complement Assumption $\ref{assum:standard}$ with assumptions on the missingness mechanism in the MCCAR setting:
\assumptionsMCCAR
Under Assumption \ref{assum:mccar}, missingness is conditionally independent of treatment given observed covariates and all individuals have some probability of treatment observation. In practice, however, missingness may also be dependent on the observed outcomes $Y$. 
Outcome-dependent missingness may occur if some records are lost or fail to be matched, or when survey non-response is present.
In this case, we say treatment is \textit{Missing at Random} (MAR) 
as missingness depends on only fully observed variables \citep{rubin_bayesian_1978, tsiatis_semiparametric_2006}. 
For example, in a study of the effect of maternal BMI on infant birthweight, the rich set of observed covariates and outcomes makes it plausible that the substantial missingness in BMI is explained by observed factors alone \citep{zhang_causal_2016}.
\citet{williamson_doubly_2012}, \citet{zhang_causal_2016}, and  
\citet{kennedy_efficient_2020} all previously studied ATE estimation in this setting under the standard Assumptions $\ref{assum:standard}$ in addition to the MAR assumption:
\assumptionsMAR

Assumption~\ref{assum:mar} excludes \textit{Missing Not at Random} (MNAR) treatment, which would occur if missingness were dependent on unobserved factors associated with treatment status itself even after conditioning on $\vX$ and $Y$.
For example, MNAR missingness may occur if participants who were less likely to receive treatment for reasons not captured by recorded covariates were also less likely to have their treatment status documented.
As treatment effects under MNAR are nonidentifiable without further assumptions, we restrict attention to MCCAR and MAR going forward.

As Fig. \ref{fig:dag} shows, in the MAR setting missingness can depend on the outcome Y, whereas under MCCAR (left panel) no such path exists since missingness depends only on observed covariates.
Fig. \ref{fig:dag} also illustrates that MCCAR \textit{R-Ignorability} (\ref{assum:mccar}) implies MAR \textit{R-Ignorability} (\ref{assum:mar}). 
Therefore, MCCAR \textit{R-Ignorability} is strictly stronger than MAR \textit{R-Ignorability}.

Although MAR \textit{R-Ignorability} is strictly weaker than MCCAR \textit{R-Ignorability}, the MAR \textit{R-Positivity} condition is possibly stronger, as it requires conditioning on $Y$ in addition to the observed covariates $X$, as summarized in Table \ref{tab:vfunction}.
However, under MCCAR Assumption~\ref{assum:mccar}, MCCAR \textit{R-Ignorability} and MCCAR \textit{R-Positivity} together imply MAR \textit{R-Positivity}. Therefore, MAR Assumption \ref{assum:mar} is strictly weaker than MCCAR Assumption \ref{assum:mccar}.

Our two sets of assumptions lead to two identification strategies. 
Because MCCAR \textit{R-Ignorability} implies MAR \textit{R-Ignorability}, the MAR identification approach is valid in both settings. 
See Appendix \ref{app:identification} for identification proofs. 

\paragraph{Strategy One} Under the standard Assumptions \ref{assum:standard} and either MCCAR Assumption \ref{assum:mccar} or MAR Assumption \ref{assum:mar}, \citet{kennedy_efficient_2020} identified the ATE using the following nuisance functions:
\begin{align*}
\nflam(\vX, Y) &= P(A = d(\vX) \mid \mathbf X, Y, R = 1)
\\ \nfpi(\vX, Y) &= P(R = 1 \mid \mathbf X, Y)
\\ \nfbeta(\vX) &= \E \left[Y\nflam(\vX, Y) \mid \vX\right] 
\\ \nfgam(\vX) &= \E\left[\nflam(\vX, Y) \mid \vX\right] = P(A = d(\vX) \mid \vX).
\end{align*}
Here, $\nfgam$ and $\nflam$ represent the probability the observational treatment assignment matches the recommendation under policy $d$ with and without outcome information,  $\nfpi$ is the treatment observation probability, and $\nfbeta$ is a propensity-weighted outcome expectation.
We can identify the value function under policy $d$ as
\begin{align}
\label{eq:maroutcomeestimator}
     V(d) = \valF = \E\left[\frac{\nfbeta(\vX)}{\nfgam(\vX)}\right] = \estMAR.
\end{align}
Importantly, $\estMAR$ (\ref{eq:maroutcomeestimator}) utilizes all data by reweighting observed outcomes by probability of treatment $\nflam$. The CATE is then identified as
\begin{align}
\label{eq:marcateestimator}
\tau(\vX) = \frac{\beta_1(\vX)}{\gamma_1(\vX)} - \frac{\beta_0(\vX)}{\gamma_0(\vX)}.
\end{align}
\paragraph{Strategy Two} The MCCAR Assumption \ref{assum:mccar} permits a second identification strategy. We define $\nnu(\vX)= \E[Y | \vX, A = a, R =1]$ as the outcome function among the observed data. Then, we may write the expected utility under policy $d$ as
\begin{align}
\label{eq:mccarval}
V(d)  = \valF = \E\left[ \nnu(\vX) \right] = \estMCCAR.
\end{align}
$\estMCCAR$ (\ref{eq:mccarval}) is a complete-case estimand as only observations without missing data are used. The same strategy yields an identified quantity for the CATE:
\begin{align}
\label{eq:cccate_estimand}
\tau(\vX) &= \nu_1(\vX) - \nu_0(\vX).
\end{align}

Unlike Strategy One, this quantity is only valid under MCCAR Assumption~\ref{assum:mccar}.
The first strategy, however, is valid under \emph{both} MCCAR and MAR Assumption \ref{assum:mar}.
If the MCCAR assumption applies, both Strategy One and Strategy Two identify the same population CATE.
As we will show, meaningful differences between the two strategies emerge in estimation, particularly in terms of efficiency and robustness to nuisance function misspecification.
We now proceed to deriving efficient and consistent estimators of these identified quantities.

\section{Efficient Policy Learning with Missing Treatments}
\label{sec:efficientpolicylearning}
\begin{table}[t]
\centering
\renewcommand{\arraystretch}{1.3}
\begin{tabular}{c|c|c}
 & MCCAR & MAR  \\
\hline
R-Ignorability & Stronger & Weaker \\
\hline
R-Positivity & Weaker & Stronger$^*$ \\
\hline
Valid IFs & $\ifMCCAR$, $\ifMAR$  &  $\ifMAR$ 
\end{tabular}
\caption{Comparing MCCAR (\ref{assum:mccar}) and MAR (\ref{assum:mar}) assumptions. 
$\mathbf{^*}$While MAR \textit{R-Positivity} is stronger, MCCAR \textit{R-Ignorability} and \textit{R-Positivity} together imply \textit{R-Positivity} in the MCCAR setting ($\nfpi(\vX_, Y) = \nfpi(\vX))$.
And so, MAR is strictly weaker.
}
\label{tab:vfunction}
\end{table}

Our goal is to learn optimal treatment policies $d^*$ that maximize expected outcomes subject to budget constraints. This requires efficient methods for both policy evaluation and treatment effect estimation.

To derive our estimators, we use Influence Functions (IFs) (also known as Neyman Orthogonal scores), which yield bias-corrected estimators that achieve the semiparametric efficiency bound and often possess double robustness (double ML) properties \citep{van_der_laan_unified_2003, tsiatis_semiparametric_2006, kennedy_semiparametric_2023, chernozhukov_doubledebiased_2018}. Doubly robust estimators remain consistent when some model components are misspecified, which is a valuable property in real-world settings with unknown nuisance functions, and can converge faster than their nuisance estimators. 
We provide technical details and all proofs in Appendix \ref{sec:ifVerification}.

\begin{remark}
    We implicitly assume $Y$ is bounded so our estimators have finite variance. We also assume that all nuisance functions are estimated from an independent sample, a common procedure known as sample splitting, which is easily adaptable to cross-fitting.
    \citep{robins_higher_2008, zheng_asymptotic_2010, chernozhukov_doubledebiased_2018}.
\end{remark}

\subsection{Policy Value Estimation}

\begin{figure*}[t]
    \centering
    \includegraphics[width=.95\textwidth]{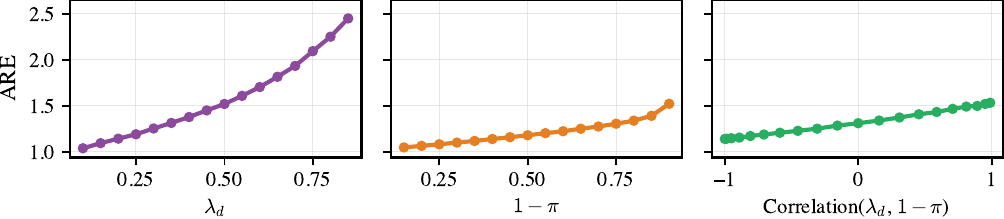}
    \caption{Asymptotic Relative Efficiency gains of $\ifMAR$ over $\ifMCCAR$ $\left(\text{ARE}(\ifMCCAR, \ifMAR) := \frac{\text{Var}(\ifMCCAR)}{\text{Var}(\ifMAR)}\right)$ as a function of the average of $\lambda_d$ (left),  the average of $1 - \pi$ (center), and the correlation between $\lambda_d$ and $1-\pi$ (right). When not varied, the average values of $\nflam$, $1 - \nfpi$, and their correlation are held fixed at 0.5. ARE increases with the augmented propensity score $\nflam$, missingness probability $1 - \nfpi$, and their correlation.}
    \label{fig:efficiency}
\end{figure*}

In this section, we establish nonparametric methods for evaluating the performance of a given policy $d$. Under each missingness assumption, we construct $\sqrt n$-consistent, asymptotically normal estimators. 
We then compare their asymptotic efficiency under MCCAR Assumption $\ref{assum:mccar}$, where both are valid. 

We consider two statistical models: $\mathcal P_{MAR}$,  all distributions on $Z$ satisfying Assumptions \ref{assum:standard} and \ref{assum:mar}, and $\mathcal P_{MCCAR}$, all distributions on $Z$ satisfying Assumptions \ref{assum:standard} and \ref{assum:mccar}. 
Under $\mathcal P_{MCCAR}$, $V(d)$ admits uncentered IFs $\ifMCCAR$ and $\ifMAR$; under $\mathcal P_{MAR}$, only $\ifMAR$ is valid:
\begin{align*}
    &\ifMAR(d) = \left(\frac{Y - \frac{\nfbeta(\vX)}{\nfgam{(\vX)}}}{\nfgam{(\vX)}}\right)\biggr(\frac{R(\1(A = d(\vX)) - \nflam(\vX))}{\nfpi(\vX, Y)}  
    \\  &\qquad \qquad \qquad  + \nflam(\vX, Y)\biggr) + \frac{\nfbeta(\vX)}{\nfgam(\vX)} \nonumber \\
    &\ifMCCAR(d) = \frac{R(\1(A = d(\vX))}{\neta(\vX)}\left(Y - \nnu(\vX)\right) + \nnu(\vX)
\end{align*}
where $\neta = P(A = d(\vX), R = 1 \mid \vX)$. We now establish convergence properties of estimators based on these IFs. 
We first present our doubly robust, nonparametric estimator for the MCCAR setting and establish its asymptotic properties:
\thmVFMCCAR
The conditions of Proposition~\ref{thm:MCCARconvergence} establish asymptotic normality under a product-rate condition on the nuisance functions and consistent estimation of $\ifMCCAR$ at any rate. 
In particular, the latter empirical process condition rules out nuisance estimators that diverge, even when the product-rate condition is satisfied.
Importantly, Proposition \ref{thm:MCCARconvergence} establishes that the estimator $\estMCCARhat$ is doubly robust, achieving consistency when either $\widehat \nnu$ or $\widehat \neta$ is correctly specified. 

In the MAR setting, $\phi_{MCCAR}$ is not a valid IF. 
We therefore adapt the IF from \citet{kennedy_efficient_2020}, originally developed for ATE estimation, to policy value estimation and derive its asymptotic variance:

\thmVFMAR

Proposition~\ref{thm:MARConvergence} demonstrates that $\estMARhat$ achieves asymptotic normality under product-rate conditions on the nuisance functions and consistent estimation of $\ifMAR$.
As in Proposition~\ref{thm:MCCARconvergence}, the empirical process condition requires consistent estimation of all nuisance functions at any rate.
Proposition~\ref{thm:MARConvergence} also reveals that consistency is achieved when $\widehat \nfgam$ is estimated consistently, along with either $\widehat \nflam$ or $\widehat \nfpi$.
Unlike standard doubly robust estimators, $\estMARhat$ requires $\widehat \nfgam$ to be consistently estimated, which is expected given that $\ifMAR$ is nonlinear in $\nfgam$.

Since $\ifMAR$ is the IF under a nonparametric model, it must be the \textit{efficient} IF under Assumption $\ref{assum:mar}$ \citep{kennedy_semiparametric_2023}. 
Unlike MAR Assumption $\ref{assum:mar}$, MCCAR Assumption $\ref{assum:mccar}$ defines a semiparametric model, which is a subset of the nonparametric model with the additional, testable condition $R \ind Y \mid X$.
Under this model, both estimators are valid, motivating a comparison of their efficiency.

For this comparison, we use the Asymptotic Relative Efficiency 
\defARE, where $f_1$ and $f_2$ are asymptotically normal estimators of the same quantity. When $\textnormal{ARE} > 1$, $f_2$ is more efficient.

\thmVarComparison

Theorem~\ref{thm:VarComparison} establishes that the complete-case-based MCCAR estimator is not the efficient choice when the conditions of both Propositions~\ref{thm:MCCARconvergence} and \ref{thm:MARConvergence} are met, demonstrating that the MAR estimator is the preferred choice under those conditions. This efficiency gain, however, is contingent upon the conditions of Proposition \ref{thm:MARConvergence}, notably that $\widehat \nfgam$ is estimated consistently, though at nonparametric rates.

As Theorem~\ref{thm:VarComparison} shows, the efficiency gain increases with the augmented propensity score ($\nflam$), missingness rate ($1-\nfpi$), and the covariance between these terms. 
To demonstrate these efficiency gains, we conduct a synthetic experiment where we systematically vary these factors. 
The experimental results in Fig. \ref{fig:efficiency} strongly confirm the predictions of Theorem~\ref{thm:VarComparison}. 
For example, ARE increases from 1.5 to 2.0 by increasing $\lambda_d$ from 45\% to $70\%$. 
Additional experimental details are provided in Appendix~\ref{app:efficiencysims}.

\subsection{Decision Rule Estimation}

With policy evaluation methods in place, we now consider the complementary challenge of learning optimal treatment assignments $d^*_\kappa(\vX)$ by estimating the CATE function $\tau(\vX)$ using the DR-Learner \citep{foster2023orthogonal, kennedy_towards_2023}.
The DR-Learner enables direct adaptation of our IF-based value function estimators to doubly-robust CATE estimation, requiring only covariates at inference time, and permits the use of generic machine learning methods for both nuisance function and CATE estimation, thus providing flexibility that approaches built on specific methods lack (such as \citet{athey2021policy, kuzmanovic_estimating_2023}).

We first define pseudo-outcomes based on the IFs of the value functions: $\widetilde{\varphi}_{\textnormal{MAR}} = \ifMARhat(1) - \ifMARhat(0)$ and $\widetilde{\phi}_{\textnormal{MCCAR}} = \ifMCCARhat(1) - \ifMCCARhat(0)$. 
When we regress these pseudo-outcomes on covariates $\vX$, the resulting CATE estimator exploits the robustness properties of the pseudo-outcome construction. 
Under additional smoothness or sparsity conditions on the CATE function, this approach can achieve faster convergence rates than methods that estimate the difference in outcome regression functions directly \citep{kennedy_towards_2023}. 
We propose two DR-Learner variants corresponding to our MAR and MCCAR assumptions:
\begin{align}
\label{eq:cateeqmar}
\tmarest &= \widehat{\mathbb E} \left[\widetilde{\varphi}_{\textnormal{MAR}} \mid \vX \right] \\
\label{eq:cateeqmccar}
\tmccarest &= \widehat{\mathbb E}\left[\widetilde{\phi}_{\textnormal{MCCAR}} \mid \vX \right]
\end{align}
While $\widehat{\tau}_{MCCAR}$ follows standard two-way sample splitting, $\widehat{\tau}_{MAR}$ requires three-way splitting for continuous $Y$ to avoid conditional density estimation when constructing $\widehat{\beta}_{d}$ and $\widehat{\gamma}_d$. 
Instead, we first fit $\widehat{\lambda}_d(\vX, Y)$. 
Using the second fold, we regress $Y\widehat{\lambda}_d(\vX, Y)$ on $\vX$ to obtain $\widehat{\beta}_{d}(\vX)$, and separately regress $\widehat{\lambda}_d(\vX, Y)$ on $\vX$ to obtain $\widehat {\gamma}_d(\vX)$.
The last fold is reserved to estimate the CATE function $\tmarest$.
Example algorithms for both estimators are detailed in Appendix~\ref{app:sec:algos}. 

Both CATE estimates can then be used directly in the budget-constrained policy $\widehat{d}^*_\kappa(\vX) = \1(\widehat{\tau}(\vX) > \widehat{\tau}^*_\kappa)$, where the threshold $\widehat{\tau}^*_\kappa$ is chosen to satisfy the budget constraint $\mathbb{E}[\hat{d}^*_\kappa(\vX)] \leq \kappa$ \citep{luedtke_optimal_2016, kitagawa_who_2018}.

With our CATE estimation strategy in place, we now verify our estimators' performance on both synthetic and semi-synthetic experiments.

\begin{remark}
Convergence rates of $\widehat \tau_{\textnormal{MAR}}$ and $\widehat\tau_{\textnormal{MCCAR}}$ can be established following the arguments of \citet{kennedy_efficient_2020}, \citet{nie_quasi-oracle_2020}, and \citet{foster2023orthogonal} using the results of Propositions \ref{thm:MCCARconvergence} and \ref{thm:MARConvergence}. 
\end{remark}
\begin{remark}
    Asymptotic normality of the estimated policy value $\widehat V$ at the data-dependent optimal policy $\widehat d^*_\kappa$ can be established under a margin condition on the CATE near threshold $\tau_{\kappa}^{^*}$ \citep{tsybakov_optimal_2004}. See Theorems~\ref{thm:policyvalueinferencemar} and \ref{thm:policyvalueinferencemccar} in Appendix~\ref{app:sec:policyvalueinference}.
\end{remark}

\section{CATE Experiments}

\begin{figure}[t]
    \centering
    \includegraphics[width=0.48\textwidth]{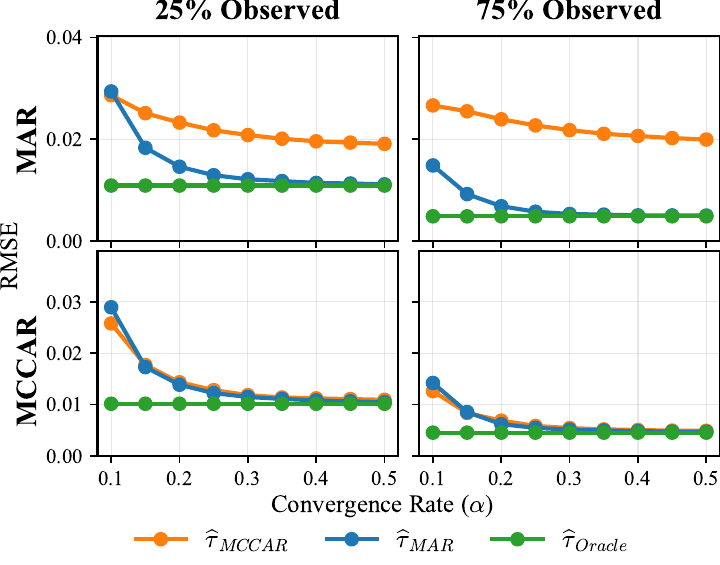}
    \caption{Median RMSE of CATE estimates comparing $\tmarest$ (blue), $\tmccarest$ (orange), and $\toracle$ (green) in the MAR (top) and MCCAR (bottom) settings over 500 iterations against the rate of noise ($\epsilon \sim N(-n^{-\alpha}, n^{-2\alpha}))$ added to all nuisance functions. In the MAR setting, $\tmccarest$ remains biased even after more treatments are observed, while in the MCCAR setting both estimators are unbiased and $\tmarest$ achieves slightly faster convergence.}
    \label{fig:cateoned}
\end{figure}

We evaluate our two CATE estimators in two experimental settings. First, we use synthetic data to verify their theoretical convergence properties. We then use semi-synthetic data from two randomized trials to evaluate policy performance.\footnote{Code for both experiments is available on Github at \href{https://github.com/jesund/learningwithmissingtreatments}{https://github.com/jesund/learningwithmissingtreatments}}.

\subsection{Synthetic Experiments}
\label{sec:syntheticexperiments}

We conduct two synthetic experiments with known potential outcomes to evaluate our estimators. 
First, we verify the robustness properties of our estimators by systematically misspecifying nuisance functions. 
Second, we evaluate their performance in a high-dimensional, sparse setting.
In both experiments, treatment and outcome rates are each 50\%.
Throughout, we use $\sigma(\cdot)$ for the expit function.
Additional implementation details are available in Appendix~\ref{app:sec:convergeneexp}.

\subsubsection{Convergence Experiment}

We create our first dataset according to the following model,
\begin{gather*}
\text{\footnotesize $X \sim$ Uni$(-1, 1)$} \\
\text{\footnotesize $Y \mid X \sim$ Ber($.45X^3 + .5$)} \\
\text{\footnotesize $A \mid X, Y \sim$ Ber($.85Y(1 - X^{\frac{2}{5}}) + .1(1-Y)X + .12$)},
\end{gather*}
which generates a smooth CATE. 
The smoothness of the resulting CATE and nuisance functions is favorable to our proposed approach.
We design two different missingness mechanisms corresponding to the MAR and MCCAR settings, 
\begin{gather*}
\text{\footnotesize R $\mid$ X, Y $\sim$ Ber($C\sigma(2YX + 2(1-X)(1-Y)) + \delta$)} \\ 
\text{\footnotesize R $\mid$ X $\sim$ Ber($C\sigma(X + 1.2 + \delta$))}
\end{gather*}
where we set $\delta=0.01$ to ensure positivity and choose $C$ to match the desired treatment observation rate. 
To test convergence properties from Propositions \ref{thm:MCCARconvergence} and \ref{thm:MARConvergence}, we add controlled noise, $\epsilon \sim \mathcal N(-n^{-\alpha}, n^{-2\alpha})$, at varying rates $\alpha$ to all nuisance functions. 
This approach allows us to precisely control the rate of convergence to evaluate how nuisance function estimation error affects CATE performance.
We compare both estimators to an oracle, using noise-free nuisance functions and estimating the CATE with smoothing splines.

\begin{figure}[t]
    \centering
    \includegraphics[width=0.48\textwidth]{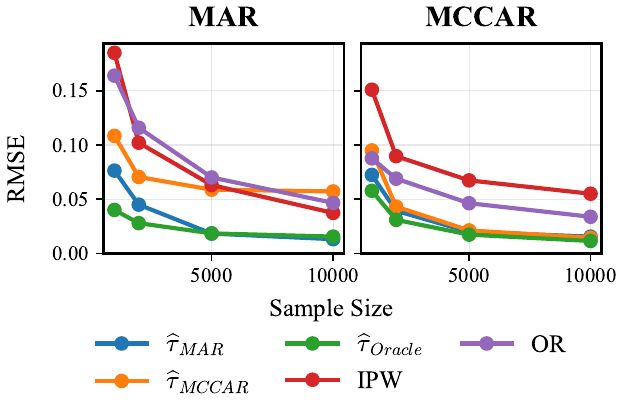}
    \caption{Average RMSE of CATE estimates comparing $\tmarest$ (blue), $\tmccarest$ (orange), $\toracle$ (green), outcome (purple) and propensity-score (brown) models in the MAR and MCCAR settings with increasing sample size. In the MAR setting, $\tmccarest$ remains biased, while in the MCCAR setting both $\tmccarest$ and $\tmarest$ converge to the oracle.}
    \label{fig:highdcate}
\end{figure}

Our findings confirm our theoretical predictions. 
As Fig. \ref{fig:cateoned} reveals, $\tmccarest$ displays persistent bias in the MAR setting, confirming it is misspecified.
Both estimators converge to the oracle in the MCCAR setting, as expected.
At very slow convergence rates, corresponding to poor nuisance function estimation, $\tmarest$ has higher error than  $\tmccarest$, but converges more quickly to the oracle.
While the observed treatment proportion does not affect asymptotic convergence rates, it does affect finite-sample performance through the asymptotic variance in 
Propositions~\ref{thm:MCCARconvergence} and~\ref{thm:MARConvergence}. 
Intuitively, more missing data inflates the inverse-probability weights and thus the variance, and so the oracle RMSE is lower at higher observation rates.
The MAR estimator partially mitigates this through the imputation model $\lambda_d$.  

\begin{figure}[t]
    \centering
    \includegraphics[width=0.48\textwidth]{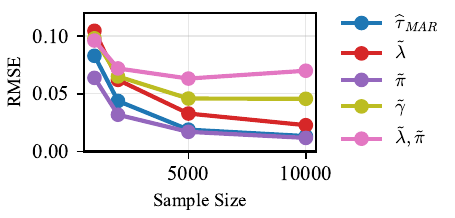}
    \caption{Average RMSE of the MAR CATE estimator under misspecification, where $\tilde f$ denotes a misspecified $f$. Under Proposition~\ref{thm:MARConvergence}, the model is robust to misspecification of $\lambda_d$ or $\pi$, but not both $\lambda_d$ and $\pi$ or $\gamma_d$.}
    \label{fig:highdimmisspec}
\end{figure}

\subsubsection{High-Dimensional Experiment}
\label{sec:highdimexp}
Our second experiment evaluates performance in a sparse setting with different CATE structure. 
In addition to oracle models, we compare our estimators to outcome regressions based on Eqs.~\ref{eq:marcateestimator} and~\ref{eq:cccate_estimand} and propensity score models appropriate for each setting.
We consider a setting with $d = 50$ covariates, where $\vX \sim N(0, I_d)$ and
\begin{gather*}
\text{\footnotesize $Y \mid \vX \sim$ Ber$(.8\sigma(3X_1 - \1(X_4 < 0) + .75\1(X_3 >.5)) + .1)$} \\
\text{\footnotesize $A \mid \vX, Y \sim$ Ber$(.7Y\1(X_3 < .5) + .2\1(-1.2 < X_{4}) + .1)$},
\end{gather*}
with missingness mechanisms for the MAR and MCCAR settings respectively:
\begin{gather*}
\text{\footnotesize $R \mid \vX, Y \sim$ Ber$(0.1\1(X_1 > 0) + 0.7 Y + 0.1)$} \\
\text{\footnotesize $R \mid \vX \sim$ Ber$(0.1\1(X_1 > 0) + 0.7\1(X_3 < 0) + 0.1),$} 
\end{gather*}
and so roughly half the units are observed, half are treated, and half have $Y=1$.
We vary $N \in \{1000, 2000, 5000, 10,000\}$ over 50 iterations using Random Forests for nuisance function and CATE estimation \citep{breiman_random_2001}. 

Our results confirm that in the MAR setting, the MCCAR estimator can be extremely biased. 
In Fig. \ref{fig:highdcate}, $\tmccarest$ RMSE remains large even as $N$ increases, highlighting that model misspecification cannot be overcome simply by collecting more data.
In contrast, both estimators attain similar error in the MCCAR setting, with neither estimator clearly outperforming the other at large sample sizes. 

Lastly, we test the robustness implications of Propositions~\ref{thm:MARConvergence} and~\ref{thm:MCCARconvergence} by systematically misspecifying nuisance functions using Random Forests with limited depth. 
Fig.~\ref{fig:highdimmisspec} validates the robustness properties of our MAR estimator. 
The estimator remains consistent when either $\lambda_d$ or $\pi$ is misspecified, but not when both are misspecified simultaneously or when $\gamma_d$ is misspecified.
Similarly, the MCCAR estimator is robust to misspecification of $\eta_d$ or $\nu_d$, but not both together.
Complete results for the MCCAR estimator are in Appendix~\ref{app:sec:convergeneexp}.

\subsection{Semi-Synthetic Experiments}
\label{sec:semisyntheticexperiments}

To evaluate our estimators under known missingness mechanisms while preserving realistic covariate and outcome structure, we construct semi-synthetic datasets by generating synthetic missingness on treatment assignment using complete data from randomized controlled trials (RCTs) in policy-relevant domains.
This approach allows us to assess policy performance across varying missingness levels, which is infeasible with naturally occurring missingness where ground truth is unknown, using true covariates and outcomes, and has previously been used in prior works studying treatment missingness (\citep{kuzmanovic_estimating_2023}).
We provide complete preprocessing, missingness generation, and model details in Appendix \ref{app:sec:semisynthetic}.

\paragraph{Datasets}
We use two datasets varying in size, covariates, and outcome type.
\textbf{Voting:} A large-scale field experiment  of 230,000 individuals studying social pressure effects on voter turnout \citep{gerber_social_2008}. We estimate the CATE of a voting history mailer on binary voting outcomes.
\textbf{STAR:} A class-size experiment with 7,000 students randomly assigned to small, regular, or regular-with-aide classes \citep{achilles_tennessees_2008}. We estimate the CATE of small vs. regular class size on kindergarten test scores.
For both datasets, we create treatment missingness under MAR and MCCAR assumptions and vary the observed treatment proportion from 10\% to 90\%. 

\paragraph{Methods} We compare our proposed estimators against several baselines spanning outcome regression, propensity weighting, and deep learning approaches.
We estimate the CATE with our proposed \textbf{DR-MAR} (Eq. \ref{eq:cateeqmar}) and \textbf{DR-MCCAR} (Eq. \ref{eq:cateeqmccar}) estimators accounting for MAR and MCCAR missingness, respectively, \textbf{OR} outcome regression models based on Eq.~\ref{eq:marcateestimator} in the MAR experiment and Eq.~\ref{eq:cccate_estimand} in the MCCAR experiment, \textbf{PW-Learner} which regresses IPW-based pseudo scores for MAR and MCCAR estimators on observed covariates (see Appendix for identification and implementation) \citep{curth_nonparametric_2021},  \textbf{MTRNET}, the adversarial learning approach of \citet{kuzmanovic_estimating_2023}. 
In Appendix~\ref{app:sec:semisyntheticresults}, we compare our models to Risk models, which directly model outcomes without using treatment and are common in policy applications \citep{sharma2025comparing}, and the standard doubly robust estimator using complete-cases only.
We tune MTRNET via Bayesian hyperparameter search \citep{akiba2019optuna}.
For all other models, we use ensembles of tuned linear models, k-Nearest Neighbors, Random Forests, and XGBoost \citep{breiman_random_2001, chen_xgboost_2016}.

\paragraph{Evaluation} We exploit the random assignment from the original experiments. 
While we introduce missingness during model training, we evaluate on the complete dataset with true treatment assignments. 
We rank individuals by their estimated treatment benefit and assign treatment to the top $\kappa-$ fraction under budget constraint $\kappa$.
We calculate policy values on a held-out test set, $V(\hat d_{\kappa}) = \mathbb E \left[Y\1(A = \hat d_{\kappa}(\vX))\right]$, under decision $d$ and budget constraint $\kappa$. 
To summarize performance across budgets, we use the Area-Under-the-Prescriptive-Effect-Curve (AUPEC) \citep{imai_experimental_2023}, which measures average policy value relative to random assignment. 
A higher AUPEC score indicates a policy has achieved higher utility, and any score over zero denotes gains over a random policy.
We evaluate all models across three sources of randomness: treatment missingness, model seeds, and train-test splits, fixing two and varying the third over 10 seeds (STAR) and 5 seeds (Voting). 
We report results varying treatment missingness here and defer other results to Appendix \ref{app:sec:semisyntheticresults}.

\begin{figure}[t]
    \centering
    \includegraphics[width=0.48\textwidth]{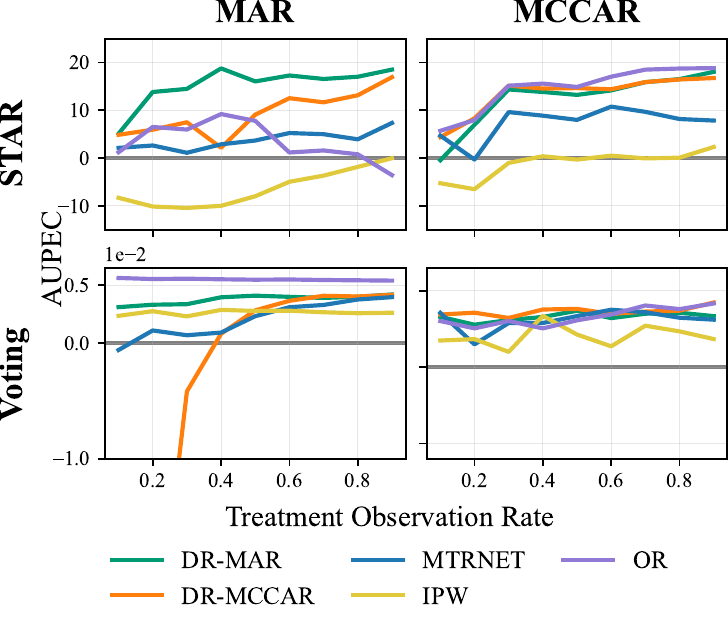}
    \caption{Average Area-Under-the-Prescriptive-Effect-Curve (AUPEC) of CATE estimators (zero is no gain over a random policy), averaged over 10 seeds (STAR) and 5 seeds (Voting) under MAR and MCCAR missingness. In the MAR setting, DR-MAR and OR, which use imputation models for missing treatment data, outperform all other methods. Under MCCAR, most methods perform similarly.}
    \label{fig:aupec}
\end{figure}

\paragraph{Results} 
MAR-based estimators demonstrate robust performance across varying treatment missingness levels and budget constraints in both the MCCAR and MAR settings, particularly when treatment data is severely limited.
Even with only 10-30\% of treatment data observed, the MAR-based OR and DR-MAR estimators maintain positive AUPEC clearly outperforming the random assignment baseline.
On the Voting dataset, the OR outperforms the DR-MAR estimator even at low levels of treatment observation. 
The Voting dataset consists predominantly of binary variables that are strongly predictive of outcomes, and so the outcome and treatment assignments models may be estimated well in this easier predictive setting, diminishing the relative advantage of the DR-MAR estimator.
On the other hand, the STAR dataset is smaller and more heterogeneous, making the treatment assignment model harder to estimate reliably.
Since the OR estimator in the MAR setting depends on this model, estimation error here can degrade its performance relative to DR-MAR.
Additionally, in the MAR setting, the DR-MCCAR estimator achieves similar performance to DR-MAR at high levels of treatment observation. 
While the DR-MCCAR estimator may be biased under MAR, the resulting policies can remain effective provided the sign and relative ordering of treatment effects are preserved.
In the MCCAR setting, most estimators perform well, with the exception of MTRNET on the STAR dataset.
We posit that this result is in part due to the smaller  ($<10k$ observations), more heterogeneous STAR dataset, which is representative of the type of setting where tree-based models typically outperform neural networks \citep{grinsztajn2022tree}.
In general, the IPW-based estimators do not perform well on our datasets under our generated missingness mechanisms.
Overall, our results confirm both the importance of flexible estimation methods and the value of properly modeling the missingness mechanism rather than discarding incomplete observations.

\section{Discussion}

In many settings with missing data, complete-case analysis is the default practice, and yet we show it is not the efficient choice even when complete-case assumptions hold.
We prove that MAR-based estimators achieve lower asymptotic variance under MCCAR assumptions.
Our experiments demonstrate that estimators constructed using MAR assumptions substantially outperform alternatives when missingness depends on outcomes, while also remaining competitive under MCCAR assumptions.

Several limitations and directions for future work merit discussion.
First, the MAR assumption may be violated in practice if unobserved factors simultaneously influence treatment assignment and the likelihood of treatment observation.
This is the MNAR setting.
Second, while our \textit{R-Positivity} assumption is necessary for identification and estimation in both the MAR and MCCAR settings, it may be violated when certain subpopulations systematically lack treatment records.
Third, the efficiency gain of the MAR estimator is an asymptotic result. 
In particular, it requires the nuisance estimators to satisfy the convergence rate conditions of Proposition~\ref{thm:MARConvergence}, which may not hold with very small sample sizes.
Finally, our results depend on accurate CATE estimation for the entire population, and so if treatment budgets are known in advance, methods that optimize directly for accurate ranking near the treatment threshold may prove more effective.
Despite these limitations, the choice between MAR and MCCAR estimators is asymmetric. 
Using MCCAR estimators under MAR assumptions causes persistent bias, while using MAR estimators under MCCAR assumptions can achieve higher efficiency.
This asymmetry reinforces the practical case for defaulting to MAR-based estimation when the missingness mechanism is uncertain.

Overall, our results have important implications for personalized decision-making across domains with missing data. 
We prove that MAR-based estimators are more efficient and require fewer identifying assumptions than MCCAR estimators, which are misspecified when missingness is also dependent on outcomes. 
Our efficient, robust, and nonparametric estimators make this recommendation practical, requiring only standard machine learning tools.
Our work enables unbiased CATE estimation in settings where complete-case approaches fail, extending personalized recommendations to the many settings where treatment data is missing.

\begin{acknowledgements} 
    EK was supported by NSF CAREER Award 2047444.
\end{acknowledgements}

\bibliography{references}

\newpage

\onecolumn

\title{Learning Who to Treat When Treatment is Missing\\(Supplementary Material)}
\maketitle
\appendix

\section{Proofs of Main Results}

\subsection{Identification Proofs (Section \ref{sec:problemMissingTreatment})}
\label{app:identification}

In this section, we identify the causal quantity of interest by linking our estimand to the observed data. We present the identification results under both our MAR and MCCAR assumptions on the missingness mechanism.
\paragraph{MAR Setting:}
Under the Standard Assumptions \ref{assum:standard} and MAR Assumptions \ref{assum:mar}, \citet{kennedy_efficient_2020} identified the ATE under treatment $A$, which we can easily use to identify the value under policy $d$:
\begin{align*}
\valF &= \int_{\mathcal X} \E \left[Y| \vX = \vx, A = d(\vX) \right] d \cP(\vx) 
\\ &= \int_{\mathcal X} \int_{\mathcal Y} \frac{y \cP(A = d(\vX) \mid \vX = \vx, Y = y)}{\cP(A = d(\vX) \mid \vX = \vx)}d\cP(Y \leq y \mid \vX =\vx)d\cP(\vx)
\\ &= \int_{\mathcal X} \int_{\mathcal Y} \frac{y \nflam(\vx, y)}{\int_{\mathcal Y} \nflam(\vx, y) d\cP(Y \leq y \mid \vX =\vx)} d\cP(Y \leq y \mid \vX = \vx)d\cP(\vx)
\\&=\idMARvf = \estMAR,
\end{align*}
where the second equality follows by Bayes' rule and third by our exchangeability assumptions. 
The identification for the CATE follows by the same argument and dropping the outer expectation:
\begin{equation*}
\label{eq:MarCateID}
\CATE = \idMARcate.
\end{equation*}
Notably, both $\nfbeta$ and $\nfgam$ can be estimated using the full sample, including observations with missing treatments. 
\paragraph{MCCAR Setting:}
Using the Standard Assumptions \ref{assum:standard} and MCCAR Assumptions \ref{assum:mccar}, we identify the value under policy $d$:
\begin{align*}
    \valF &= \int_{\mathcal X} \E \left[Y| \vX = \vx, A = d(\vX)\right] d\cP(\vx) \\ 
    &= \int_{\mathcal X} \E \left[Y| \vX = \vx, A = d(\vX), R = 1\right] d \cP(\vx) \\ 
    &= \idMCCARvf = \estMCCAR.
\end{align*}
And so, the CATE is identified as:
\begin{equation*}
\label{eq:MccarCateId}
\CATE = \idMCCARcate.
\end{equation*}
We remark that under Assumption $\ref{assum:mccar}$ the value of policy $d$ is still identified by $\idMARvf$. 
This result follows by the same proof as the MAR setting and noting that the MCCAR \textit{R-ignorability} assumption, $\asMCCAR$ implies the weaker MAR \textit{R-ignorability} assumption, $\asMAR$.

\subsection{Proofs of IF Results (from Section \ref{sec:efficientpolicylearning})}
\label{sec:ifVerification}

In this section, we prove Propositions \ref{thm:MCCARconvergence} and \ref{thm:MARConvergence}. We then use these results to prove Theorem \ref{thm:VarComparison}. 
Note that while Section \ref{sec:efficientpolicylearning} presents the uncentered IFs $\ifMCCAR$ and $\ifMAR$ used in constructing our estimators, in the proofs in this section we mainly use the centered IFs: 
\begin{align*}
\IFMCCAR(z, \Ps) &= \ifMCCAR(z, \Ps) - \estMCCAR \\
\IFMAR(z, \Ps) &= \ifMAR(z, \Ps) - \estMAR. 
\end{align*}
For each proposition, we verify our candidate IFs by the von Mises Expansion (also known as the distributional Taylor Expansion):
\begin{equation}
\label{eq:vmexpansion}
\vmExpansion
\end{equation}
where $\psi$ is our estimand of interest for population $P$, $\overline P$ is a perturbed distribution on $P$, influence function $\IFu$ is a mean-zero function with finite variance, and $R_2(\overline P, P)$ is a second-order error term. 
As a result of this second-order remainder, functions $\IFu$ which satisfy this expansion achieve faster convergence to the true estimand $\psi$ and under nonparametric statistical models also achieve the semiparametric efficiency bound \citep{kennedy_semiparametric_2023}. 
Using the results of the von Mises Expansion, we can construct doubly robust estimators and analyze the asymptotic properties of these estimators:

\begin{lemma}
\label{lemma:ifestimators}
Let $\IFu(z, \Ps) = \varphi(z, \Ps) - \Pn(\psi(\Ps))$ satisfy  Eq. \ref{eq:vmexpansion} for estimand $\psi$ under estimated distribution $\Ps$. Then, we may define the bias-corrected estimator as:
\begin{align*}
\widehat \psi =  \Pn(\IFu(Z, \Ps)) + \psi(\Ps) = \Pn(\varphi(Z, \Ps)).
\end{align*}
Assume that all nuisance functions are estimated from an independent sample and that $\|\IFu(z, \Ps) - \IFu(z, \mathbb P)\| = \lilo(1)$. Then, by Proposition 1 from \citet{kennedy_semiparametric_2023}, we have:
\begin{align*}
\widehat \psi - \psi &=  \Pn(\IFu(Z, \Ps)) + \psi(\Ps) - \psi(\mathbb P) \\ 
&= \Pn(\IFu(Z, \mathbb P)) - \mathbb E \left[\IFu(Z; \mathbb P)\right] + R_2(\Ps, \mathbb P) +\lilo\left(1/\sqrt{n}\right).
\end{align*}
The first two terms are the difference of a sample average and the true population mean. If $R_2(\Ps, \mathbb P) = \lilo\left(1 /\sqrt n\right)$, then by the Central Limit Theorem and Slutsky's Theorem,
\begin{align*}
\sqrt n(\widehat \psi - \psi) \rightsquigarrow N\left(0, \var(\IFu(Z; \mathbb P))\right).
\end{align*}
\end{lemma}
Therefore, to establish $\sqrt{n}$-consistency and asymptotic normality under Lemma \ref{lemma:ifestimators}, it suffices to provide conditions under which $R_2(\Ps, \mathbb P) = o_{\mathbb{P}}(1/\sqrt{n})$ for an IF satisfying the von Mises Expansion (Eq. \ref{eq:vmexpansion}).

\subsubsection{Proof of Proposition \ref{thm:MCCARconvergence}}

Recall our bias-corrected estimator $\estMCCARhat = \Pn(\IFMCCAR(Z, \Ps)) + \estMCCAR(\Ps) = \Pn(\ifMCCAR(Z, \Ps))$.
We first verify that $\IFMCCAR$ satisfies the von Mises Expansion (Eq. \ref{eq:vmexpansion}) and derive the conditions under which the remainder term satisfies $R_2(\Ps, \mathbb P) = o_{\mathbb{P}}(1/\sqrt{n})$. 
We conclude by deriving the asymptotic variance.

\paragraph{Verifying the IF and Deriving the Remainder Term}
Recall our candidate influence function for the value function in the MCCAR setting:
\begin{align}
\label{eq:MccarIf}
\IFMCCAR  = \eqIfMCCAR.
\end{align}
In this section, we will prove that the candidate IF (Eq. \ref{eq:MccarIf}) admits a von Mises Expansion (Eq. \ref{eq:vmexpansion}) with second-order remainder bounded by
\begin{equation*}
R_2(\Ps, \cP) \lesssim \|\neta - \widehat \neta \|\|\nnu - \widehat \nnu \|,
\end{equation*}
where we take the estimated distribution $\Ps$ to be the perturbed distribution $\overline P$. 
We begin by showing that $\IFMCCAR$ (Eq. \ref{eq:MccarIf}) is mean zero:
\begin{align*}
\E \left[\IFMCCAR \right] &= \E \left[\frac{R\1(A = d(\vX))(Y - \nnu(\vX))}{\neta(\vX)} + \nnu(X)\right]  - \E \left[\nnu(\vX)\right] \\ 
&= \E \left[\frac{\neta(\vX)}{\neta(\vX)}\biggr( \E \left[Y\mid \vX = \vx, A = d(\vX), R =1 \right] - \nnu(\vX) \biggr)\right] = 0,
\end{align*}
where the second equality follows by iterated expectation over $\vX$ and the fact that $R, \1(A = d(\vX))$ are binary.
Using this result, we may arrange the von Mises Expansion (Eq. \ref{eq:vmexpansion}) in terms of the remainder
\begin{equation}\label{app:eqmccarremainder}
R_2(\Ps, \cP) 
= \estMCCAR (\Ps) - \estMCCAR (\cP) + \int \IFMCCAR(z; \Ps)d \cP (z).
\end{equation}
Expanding the final term of the remainder (\ref{app:eqmccarremainder}),
\begin{align*}
\int \IFMCCAR(z; \Ps)d\cP(z) &= 
\int \biggr\{ \frac{R \mathbbm 1(A = d(\vx))(Y - \widehat \nnu(\vx))}{\widehat \neta(\vx)} + \widehat \nnu(\vx) - \estMCCAR(\Ps)\biggr\}d\cP(z)
\\ &= \int \biggr\{\frac{\neta(\vx)}{\widehat \neta(\vx)} \cdot (\nnu(\vx) - \widehat \nnu(\vx)) + \widehat \nnu(\vx)\biggr\}d\cP(z) - \estMCCAR(\Ps) 
\\ &= \int \biggr\{\frac{\neta(\vx)}{\widehat \neta(\vx)} \cdot (\nnu(\vx) - \widehat \nnu(\vx)) - (\nnu(\vx)  - \widehat \nnu(\vx))\biggr\}d\cP(z) + \estMCCAR(\cP) - \estMCCAR(\Ps)  
\\ &= \int \left(\frac{1}{\widehat \neta(\vx)} - \frac{1}{\neta(\vx)}\right)(\nnu(\vx) - \widehat \nnu(\vx))(\neta(\vx))d\cP(z) + \estMCCAR(\cP) - \estMCCAR(\Ps)  
\end{align*}
where the second equality follows by iterated expectation over $\vX$ and the fact that $R, \1(A = d(\vX))$ are binary. And thus, the remainder (\ref{app:eqmccarremainder}) is equivalently
\begin{equation*}
R_2(\Ps, \cP) =  \int \left(\frac{1}{\widehat \neta(\vx)} - \frac{1}{\neta(\vx)}\right)(\nnu(\vx) - \widehat \nnu(\vx))(\neta(\vx))d\cP(\vx).
\end{equation*}
Under $A-$ and $R-$positivity (Assumptions \ref{assum:standard} and \ref{assum:mccar}), we have
\begin{align*}
|R_2(\Ps, \cP)| \leq \frac{1}{\epsilon} \int | \neta(\vx) - \widehat \neta(\vx)| |\nnu(\vx) - \widehat \nnu(\vx)|d\cP(\vx) \lesssim \| \neta(\vx) - \widehat \neta(\vx) \| \| \nnu(\vx) - \widehat \nnu(\vx)\|,
\end{align*}
where the final inequality follows by Cauchy-Schwarz.
As a result, the remainder term is second-order and we conclude that the von Mises Expansion (Eq. \ref{eq:vmexpansion}) holds. 

And so, Lemma \ref{lemma:ifestimators} applies when $\| \neta(\vx) - \widehat \neta(\vx) \| \| \nnu(\vx) - \widehat \nnu(\vx)\| = \lilo(1/\sqrt n)$, establishing the convergence properties of Proposition \ref{thm:MCCARconvergence}. 

\paragraph{Asymptotic Variance}

To complete the proof of Proposition \ref{thm:MCCARconvergence}, we derive the asymptotic variance.

Using the fact that $\E \left[\IFMCCAR\right] = 0$ (shown above), 
\begin{align}
\label{app:eq:varMCCARlong}
\var(\ifMCCAR) &= \E \left[\left(\eqIfMCCAR\right)^{2}\right] \nonumber
\\&= \E \left[\left(\frac{R\1(A = d(\vX))(Y-\nnu(\vX))}{\neta(\vX)}\right)^{2}\right] + \E \left[\left( \nnu(\vX) - \estMCCAR \right)^{2}\right]  \nonumber
\\&\quad +  \E \left[ \left(\frac{R\1(A = d(\vX))(Y-\nnu(\vX))}{\neta(\vX)}\right) \cdot \left( \nnu(\vX) - \estMCCAR \right)\right].
\end{align}
We begin by showing that the final term of Eq. \ref{app:eq:varMCCARlong} is zero. By iterated expectation,
\begin{align*}
&\E \left[\E \left[\left(\frac{R\1(A = d(\vX))(Y - \nnu(\vX))}{\neta(\vX)} \right) \left(\nnu(\vX) - \estMCCAR \right) \mid \vX\right]\right] \\ 
&= \E \left[\frac{\neta(\vX)}{\neta(\vX)} \left(\E \left[Y|\vX, R = 1, A = d(\vX) \right] - \nnu(\vX)\right)(\nnu(\vX) - \estMCCAR)\right] = 0,
\end{align*} 
which holds because $R, \1(A = d(\vX))$ are binary. Then, rewriting the first term of the Eq. \ref{app:eq:varMCCARlong},
\begin{align}
\label{app:eq:varstep}
\E \left[\left(\frac{R\1(A = d(\vX))(Y-\nnu(\vX))}{\neta(\vX)}\right)^{2}\right] &= \E \left[\frac{R\1(A = d(\vX))(Y - \nnu(\vX))^{2}}{\neta(\vX)^{2}}\right] \equiv \E \left[\frac{\sigma_{d}^{2}(\vX)}{\neta(\vX)}\right]
\end{align}
where the second equality follows by iterated expectation. It follows that
\begin{equation}\label{app:eq:varMCCAR}
\var(\ifMCCAR) = \eqVarMCCAR.
\end{equation}

\paragraph{Deriving the Candidate IF}

For completeness, we include our derivation of the candidate IF. We follow the "tricks" from \citet{kennedy_semiparametric_2023}, namely (1) treating the data as discrete, (2) treating the IF as a derivative, and (3) utilizing known IFs, such as the IF of a conditional expectation or density. 
First, following \citet{kennedy_semiparametric_2023}, we define $\text{InfFun} : \psi \to L_2(\mathbb P)$ as an operator that maps a function $\psi$ to its influence function under a nonparametric model. Then,
\begin{align*}
\text{InfFun}(\estMCCAR) &= \text{InfFun}\left(\sum_{x} \nu_d(x)p(x)\right) \\ 
&= \sum_{x}\big(\text{InfFun}(\nu_d(\vx))p(\vx) + \nu_d(\vx)\text{InfFun}(p(\vx))\big) \\ 
&= \sum_{x}\left( \frac{\1(\vX = \vx, R = 1, A = d(\vx))}{p(\vX = \vx, R =1 , A = d(\vx))}\left(Y - \nu_d(\vx)\right)p(\vx) + \nu_d(\vx)(\1(\vX = \vx) - p(\vx)) \right) \\ 
&= \frac{R\1(A = d(\vX))}{\eta_d(\vX)}(Y - \eta_d(\vx)) + \nu_d(\vX) - \estMCCAR,
\end{align*}
which is exactly the centered influence function in (\ref{eq:MccarIf}).

\subsubsection{Proof of Proposition \ref{thm:MARConvergence}}

Recall our bias-corrected estimator $\estMARhat = \Pn(\IFMAR(Z, \Ps)) + \estMAR(\Ps) = \Pn(\ifMAR(Z, \Ps))$.
We now verify that $\ifMAR$ satisfies the von Mises Expansion (Eq. \ref{eq:vmexpansion}) and provide sufficient conditions on the nuisance functions for $\sqrt{n}$-consistency and asymptotic normality by analyzing the remainder term and applying Lemma \ref{lemma:ifestimators}. 
We then derive the asymptotic variance.

\paragraph{Verifying the IF and Deriving the Remainder Term}

\cite{kennedy_efficient_2020} verified the influence function for the average effect of treatment $a$ under MAR assumptions.
We first note that the effect under policy $d$ can be seen as a generalization as the average effect of treatment $a$, as this can be represented by a policy that gives all individuals treatment $a$. 
Therefore, using this fact, the remainder term Lemma 1 of \citet{kennedy_efficient_2020} may be written as:
\begin{align*}
R_2(\Ps, \mathbb P) &= \int \left(\frac{y - \frac{\widehat \nfbeta(\vx)}{\widehat \nfgam(\vx)}}{\widehat \nfgam(\vx)}\right)\left(\frac{\pi(\vx, y) - \widehat \pi(\vx, y)}{\widehat \nfpi(\vx, y)}\right)\left(\nflam(\vx, y) - \widehat \nflam(\vx, y)\right) \\ 
&\qquad + \left(\frac{\nfbeta(\vx)}{\nfgam(\vx)} - \frac{\widehat \nfbeta(\vx)}{\widehat \nfgam(\vx)}\right)\left(\frac{\nfgam(\vx) - \widehat \nfgam(\vx)}{\widehat \nfgam(\vx)}\right)d\mathbb P(z)
\end{align*}

And so, the remainder term is second-order and we conclude that the von Mises Expansion (Eq. \ref{eq:vmexpansion}) holds. 
As \citet{kennedy_efficient_2020} noted, the conditions of Lemma \ref{lemma:ifestimators} are met when $\|\nfpi - \widehat \nfpi \|\|\nflam -\widehat \nflam\|  + \|\nfbeta - \widehat \nfbeta \|\|\nfgam -\widehat \nfgam\| + \|\nfgam - \widehat \nfgam\|^{2} = \lilo(1/\sqrt n)$.

\paragraph{Asymptotic Variance}

To complete the proof of Proposition \ref{thm:MARConvergence}, we derive the asymptotic variance.

Because $\E \left[\IFMAR\right] = 0$, $\var(\ifMAR) = $ 
\begin{align}
\label{app:eq:MARvarexpanded}
&\E \left[\left(\eqIfMAR\right)^{2}\right] \nonumber
\\ &= \E \left[\biggr\{\left(\frac{Y - \nfbeta(\vX) / \nfgam(\vX) }{\nfgam(\vX)}\right) \left(\frac{R \left(\1(A = d(\vX)) - \nflam(\vX,Y) \right)}{\nfpi(\vX,Y)} + \nflam (\vX,Y) \right)\biggr\}^{2}\right] \nonumber
\\&\quad+ \E \left[\left(\frac{\nfbeta(\vX)}{\nfgam(\vX)} - \estMAR\right)^{2}\right]  \nonumber
\\&\quad + \E \left[\left(\frac{Y - \nfbeta(\vX) / \nfgam(\vX) }{\nfgam(\vX)}\right) \left(\frac{R \left(\1(A = d(\vX)) - \nflam(\vX,Y) \right)}{\nfpi(\vX,Y)} + \nflam (\vX,Y) \right)\left(\frac{\nfbeta(\vX)}{\nfgam(\vX)} - \estMAR\right)\right].
\end{align}
We first show that the cross-product term in Eq. \ref{app:eq:MARvarexpanded} is zero. By iterated expectation over $\vX$ and $Y$,
\begin{align*}
&\E \left[ \E \left[\left(\frac{Y - \nfbeta(\vX) / \nfgam(\vX) }{\nfgam(\vX)}\right) \left(\frac{R \left(\1(A = d(\vX)) - \nflam(\vX,Y) \right)}{\nfpi(\vX,Y)} + \nflam (\vX,Y) \right)\left(\frac{\nfbeta(\vX)}{\nfgam(\vX)} - \estMAR\right) \mid \vX, Y \right] \right] 
\\ &\quad= \E \left[\left(\frac{\nfbeta(\vX)}{\nfgam(\vX)} - \estMAR\right)  \left(\frac{Y - \nfbeta(\vX) / \nfgam(\vX) }{\nfgam(\vX)}\right) \left(\frac{\nfpi(\vX, Y) \left(\nflam(\vX, Y) - \nflam(\vX,Y) \right)}{\nfpi(\vX,Y)} + \nflam (\vX,Y) \right)\right]
\\ &\quad= \E \left[\left(\frac{\nfbeta(\vX)}{\nfgam(\vX)} - \estMAR\right)  \left(\frac{\E[Y\nflam (\vX,Y) \mid \vX] - \nfbeta(\vX) / \nfgam(\vX)  \E[\nflam (\vX,Y)\mid \vX] }{\nfgam(\vX)}\right)\right] = 0.
\end{align*}
Rearranging the first-term in Eq. \ref{app:eq:MARvarexpanded},
\begin{align*}
&\E \left[\biggr\{\left(\frac{Y - \nfbeta(\vX) / \nfgam(\vX) }{\nfgam(\vX)}\right) \cdot \left(\frac{R \left(\1(A = d(\vX)) - \nflam(\vX,Y) \right)}{\nfpi(\vX,Y)} + \nflam (\vX,Y) \right)\biggr\}^{2}\right] 
\\&=\E \left[\left(\frac{Y - \nfbeta(\vX) / \nfgam(\vX) }{\nfgam(\vX)}\right)^{2} \cdot \biggr\{\left(\frac{R \left(\1(A = d(\vX)) - \nflam(\vX,Y) \right)}{\nfpi(\vX,Y)}\right)^{2} + \nflam^{2} (\vX,Y)\biggr\}\right] 
\\&=\E \left[\left(\frac{Y - \nfbeta(\vX) / \nfgam(\vX) }{\nfgam(\vX)}\right)^{2} \cdot \biggr\{\left(\frac{R \left(\1(A = d(\vX)) - 2\1(A = d(\vX))\nflam(\vX,Y) + \nflam^{2}(\vX,Y) \right)}{\nfpi^{2}(\vX,Y)}\right) + \nflam^{2} (\vX,Y)\biggr\}\right] 
\\&= \E \left[\left(\frac{Y - \nfbeta(\vX) / \nfgam(\vX) }{\nfgam(\vX)}\right)^{2} \cdot \left(\frac{\nflam(\vX,Y)- \nflam^{2}(\vX,Y)}{\nfpi(\vX,Y)} + \nflam^{2} (\vX,Y)\right)\right] 
\\&= \E \left[\left(\frac{Y - \nfbeta(\vX) / \nfgam(\vX) }{\nfgam(\vX)}\right)^{2} \cdot \left(\frac{\nflam(\vX,Y)}{\nfpi(\vX,Y)} - \nflam^{2}(\vX,Y)\cdot \biggr\{\frac{1 - \nfpi(\vX,Y)}{\nfpi(\vX,Y)}\biggr\}\right)\right] \\
&= \E \left[\left(\frac{Y - \nfbeta(\vX) / \nfgam(\vX) }{\nfgam(\vX)}\right)^{2} \cdot \left(\frac{\nflam(\vX,Y)}{\nfpi(\vX,Y)} - \nflam^{2}(\vX,Y)\cdot \biggr\{\frac{1 - \nfpi(\vX,Y)}{\nfpi(\vX,Y)}\biggr\}\right)\right]
\end{align*}
where the first equality holds because the cross term is eliminated by iterated expectation over $\vX, Y$ ($\E[\1(A = d(\vX)) |\vX, Y)] - \nflam(\vX, Y) = 0)$, and the third equality by again taking iterated expectation over $\vX$ and $Y$. Therefore, we have shown 
\begin{align}
\label{eq:varMAR}
&\var(\ifMAR)= \nonumber
\\& \eqVarMAR.
\end{align}

\begin{remark}
\label{remark:atecovariance}
We note the variance of an estimand with contrasts (such as the ATE) may have an additional term from the covariance. Using the ATE as an example, the covariance between the residual terms is:
\begin{align*}
&-2\E \biggr[\biggr\{\left(\frac{Y - \beta_1(\vX)/\gamma_1(\vX)}{\gamma_1(\vX)}\right) \left(\frac{R(\1(A = 1) - \lambda_1(\vX, Y))}{\pi(\vX, Y)} + \lambda_1(\vX, Y)\right) \biggr\} 
\\& \qquad \times  \left(\frac{Y - \beta_0(X)/\gamma_0(\vX)}{\gamma_0(\vX)}\right) \left(\frac{R(\1(A = 0) - \lambda_0(\vX, Y))}{\pi(\vX, Y)} + \lambda_0(\vX, Y)\right) \biggr\}\biggr] \\ 
&= -2\E \biggr[ \biggr\{\left(\frac{Y - \beta_1(\vX)/\gamma_1(\vX)}{\gamma_1(\vX)}\right) \left(\frac{Y - \beta_0(\vX)/\gamma_0(\vX)}{\gamma_0(\vX)}\right) \biggr\} \biggr\{ \frac{-\lambda_1(\vX, Y) \lambda_{0}(\vX, Y)}{\pi(\vX, Y)} + \lambda_{1}(\vX, Y) \lambda_{0}(\vX, Y)\biggr\} \biggr]
\\&= 2\E \left[\biggr\{\frac{1 - \pi(\vX, Y)}{\pi(\vX, Y)}\biggr\} \biggr\{\frac{(Y - \beta_1(\vX)/\gamma_1(\vX))}{\gamma_1(\vX)} \biggr\} \biggr\{\frac{(Y - \beta_0(\vX)/\gamma_0(\vX)) }{\gamma_0(\vX)} \biggr\}\biggr\{\lambda_{1}(\vX, Y)\lambda_{0}(\vX, Y)\biggr\}\right].
\end{align*}
Under complete data, the covariance of the residuals, i.e. $A(Y - \nu_1)(1-A)(Y - \nu_0)$, are zero, which as we see above does not happen with $\ifMAR$. Because of this, we may lose efficiency if the residuals have the same sign.
\end{remark}

\subsubsection{Proof of Theorem \ref{thm:VarComparison}}

As $\ifMAR$ is also a valid IF under MCCAR Assumptions $\ref{assum:mccar}$, we conclude by deriving an expression for the asymptotic relative efficiency of estimators built using $\ifMAR$ compared to $\ifMCCAR$. 

We begin by rewriting $\var(\ifMAR)$ in terms of $\var(\ifMCCAR)$.  First, we note the last terms of both expressions are equivalent:
\begin{align}
\label{app:eq:outcomevar}
\E \biggr[\biggr(\frac{\nfbeta(\vX)}{\nfgam(\vX)} - \estMAR\biggr)^{2}\biggr] = \E \left[\left( \nnu(\vX) - \estMCCAR \right)^{2}\right],
\end{align}
which follows because $\estMAR = \E \biggr[\frac{\nfbeta(\vX)}{\nfgam(\vX)}\biggr] = V(d) = \mathbb E \left[\nnu(\vX)\right] = \estMCCAR$. It is also possible to verify this equality by reversing the steps of the MAR Identification strategy (see \ref{app:identification}). Now, considering the first term of $\var(\ifMAR)$ (Eq. \ref{eq:varMAR}), 
\begin{align}
\label{app:eq:vargain}
\E \biggr[&\biggr\{\frac{Y - \nfbeta(\vX)/\nfgam(\vX)}{\nfgam(\vX)}\biggr\}^{2}\cdot 
\biggr\{\frac{\nflam(\vX, Y)}{\pi(\vX, Y)} - \nflam^2(\vX,Y)\left(\frac{1-\pi(\vX,Y)}{\pi(\vX, Y)}\right)\biggr\}\biggr] \nonumber \\ 
&= \E \biggr[\biggr\{\frac{(Y - \nnu(\vX))^{2}}{\nfgam(\vX)^{2}}\biggr\}\cdot 
\biggr\{\frac{\1(A = d(\vX)}{\pi(\vX,Y)} - \nflam^2(\vX,Y)\left(\frac{1-\pi(\vX,Y)}{\pi(\vX, Y)}\right)\biggr\}\biggr] \nonumber \\
&= \E \left[\frac{\sigma_{d}^{2}(\vX)}{\neta(\vX)}\right] - \varGain \nonumber \\ 
&\equiv \E \left[\frac{\sigma_{d}^{2}(\vX)}{\neta(\vX)}\right] - g(d, \mathbb P)
\end{align}
where the first equality follows by iterated expectation over $\vX$ and $Y$ and the fact that $A$ is binary ($\E \left[\E \left[\1(A | \vX, Y) \mid \vX, Y\right]\right]=A)$ as well as \textit{R-Ignorability} from Assumption \ref{assum:mccar}. 
The third equality also follows from \textit{R-Ignorability} of Assumption \ref{assum:mccar}. 
Therefore, by Eq. \ref{app:eq:outcomevar} and Eq. \ref{app:eq:vargain}, we may rewrite $\var(\ifMAR)$ as
\begin{align*}
\var(\ifMAR) &= \E \left[\left( \nnu(\vX) - \estMCCAR \right)^{2}\right] + \E \left[\frac{\sigma_{d}^{2}(\vX)}{\neta(\vX)}\right] - g(d, \mathbb P) 
\\&= \var(\ifMCCAR) - g(d, \cP).
\end{align*}
Using this result, and omitting the argument for the decision policy $d$ in $g$, we can derive an expression for the ARE, 
\begin{align*}
\textnormal{ARE}(\var(\ifMCCAR), \var(\ifMAR)) &= \frac{\var(\ifMCCAR)}{\var(\ifMAR)}
\\ &= \frac{\var(\ifMCCAR)}{\var(\ifMCCAR) - g(\mathbb P)} \\ &= \biggr(1 - \frac{g(\mathbb P)}{\var(\ifMCCAR)}\biggr)^{-1}.
\end{align*}
Hence, as $g(\mathbb P) \geq 0$, we conclude that estimators based on $\ifMAR$ are more efficient.

\begin{remark}

Recall that the asymptotic variance of contrasts of $\ifMAR$ will have an additional covariance term (Remark \ref{remark:atecovariance}). We now show that despite this penalty $\varphi_{MAR}$ is still more efficient. Notice 
\begin{align*}
\var(\ifMCCAR(1) - \ifMCCAR(0)) &= \E \left[\frac{\sigma_{1}^{2}(\vX)}{\eta_{1}(\vX)}\right] + \left[\frac{\sigma_{0}^{2}(\vX)}{\eta_{0}(\vX)}\right]  + \mathbb E \left[(\tau(\vX) - \tau)^{2}\right]
\end{align*}
And then using the argument above and covariance term from Remark \ref{remark:atecovariance},
\begin{align*}
&\var(\ifMAR(1) - \ifMAR(0)) = \var(\ifMCCAR(1) - \ifMCCAR(0)) - g(1,\mathbb P) - g(0, \mathbb P) 
\\&\quad +2\E \left[\biggr\{\frac{1 - \pi(\vX)}{\pi(\vX)}\biggr\} \biggr\{\frac{(Y - \nu_1)}{\gamma_1(\vX)} \biggr\} \biggr\{\frac{(Y - \nu_0) }{\gamma_0(\vX)} \biggr\}\biggr\{\lambda_{1}(\vX, Y)\lambda_{0}(\vX, Y)\biggr\}\right]
\end{align*}
If the covariance term (last term) is negative, which occurs when the product of the expectation of the residuals is negative, then $\ifMAR$ is still more efficient. 
If the covariance term is positive, we only achieve gains if $g(1, \mathbb P) + g(0, \mathbb P)$ is greater than the covariance term. Dropping arguments for notational convenience,
\begin{gather*}
g(1, \mathbb P) + g(0, \mathbb P) \geq 2\mathbb E \left[\left(\frac{1-\pi}{\pi}\right)\left( \frac{Y - \nu_1}{\gamma_1}\right)\left( \frac{Y - \nu_0}{\gamma_0}\right)\lambda_1\lambda0\right] \\ 
\mathbb E \left[\left(\frac{1-\pi}{\pi}\right)\biggr\{\left(\frac{(Y - \nu_1)\lambda_1}{\gamma_1}\right)^{2}-2\left(\frac{(Y - \nu_1)\lambda_1}{\gamma_1}\right)\left( \frac{(Y - \nu_0)\lambda_0}{\gamma_0}\right)+\left(\frac{(Y - \nu_0)\lambda_0}{\gamma_0}\right)^{2} \biggr\}\right] \geq 0
\end{gather*}
which clearly holds. And so, despite the covariance penalty in the ATE case we achieve efficiency gains.
\end{remark}

\subsection{Policy Values under Estimated Policies}
\label{app:sec:policyvalueinference}

We establish the asymptotic normality of the policy value under a fixed policy $d$ in the main text. 
In this section, we extend these results to policy value estimation at the data-dependent optimal policy $\widehat d^*_{\kappa}$. In order to obtain asymptotic normality in this case, we need to invoke an additional assumption:

\begin{assumption}[Margin Condition] \label{ass:margincondition} 
There exists constants $C > 0$ and $\alpha \geq 0$ such that 
\[
\mathbb{P}(0 < |\tau(\vx) - \tau^*_{\kappa} | \leq t) \leq Ct^\alpha \text{ for all } t > 0.
\]
\end{assumption}
The margin condition, originating from \citet{tsybakov_optimal_2004}, limits the probability mass concentrating near the optimal decision boundary $\tau^{*}_{\kappa}$. 
It is satisfied whenever the CATE distribution has a continuous density near $\tau^*_{\kappa}$, and therefore is mild in practice.

We can now use Assumption~\ref{ass:margincondition} to extend asymptotic normality to policy value estimation an estimated policy $\widehat d^{*}_{\kappa}$ in both the MAR and MCCAR settings:

\begin{theorem} \label{thm:policyvalueinferencemccar}
Define the bias-corrected estimator derived from $\ifMCCAR$ at the estimated policy $\widehat d_{\kappa}^{*}$ as $\estMCCARhat(\widehat d_{\kappa}^{*}) = \Pn(\ifMCCARhat(\widehat d_{\kappa}^{*}))$. Assume the conditions of Proposition~\ref{thm:MCCARconvergence}, in addition to Assumption~\ref{ass:margincondition} and $\|(\widehat \tau - \widehat \tau^*_{\kappa}) - (\tau - \tau^*_{\kappa}) \|^{1 + \alpha}_{\infty} = o_{\mathbb P}(n^{-1/2} )$. 
Then, $\sqrt{n}(\estMCCARhat(\widehat d_{\kappa}^{*}) - V(d^*_{\kappa})) \rightsquigarrow \mathcal{N}(0, \var(\ifMCCAR(d^*_{\kappa})))$.
\end{theorem}

\begin{theorem}
    \label{thm:policyvalueinferencemar}
    Define the bias-corrected estimator derived from $\ifMAR$ at the estimated policy $\widehat d_{\kappa}^{*}$ as $\estMARhat(\widehat d_{\kappa}^{*}) = \Pn(\ifMARhat(\widehat d_{\kappa}^{*}))$. Assume the conditions of Proposition~\ref{thm:MARConvergence}, in addition to Assumption~\ref{ass:margincondition} and $\|(\widehat \tau - \widehat \tau^*_{\kappa}) - (\tau - \tau^*_{\kappa}) \|^{1 + \alpha}_{\infty} = o_{\mathbb P}(n^{-1/2} )$. 
Then, $\sqrt{n}(\estMARhat(\widehat d_{\kappa}^{*}) - V(d^*_{\kappa})) \rightsquigarrow \mathcal{N}(0, \var(\ifMAR(d^*_{\kappa})))$.
\end{theorem}

\subsubsection{Proof of Theorem~\ref{thm:policyvalueinferencemccar}}

We state the proof for Theorem~\ref{thm:policyvalueinferencemccar}; the proof for Theorem~\ref{thm:policyvalueinferencemar} is analogous. 
In this section, we let $\mathbb P(f) \equiv \mathbb E \left[f(\vX)\right]$ for a generic function $f$. 
First, we rewrite the identified value function in the MCCAR setting as
\begin{align*}
V(d^*_{\kappa}) &= \mathbb E \left[Y(d^*_{\kappa})\right] = \mathbb E \left[\tau(\vX) \1(\tau(\vX) > \tau^*_{\kappa})] + \E[\nu_0(\vX)\right].
\end{align*}
Then, recall that we can rewrite the influence function as
\begin{align*}
\ifMCCAR(d) &= \frac{R\1(A = d)}{\neta(\vX)}\left(Y - \nnu(\vX)\right) + \nnu(\vX) \nonumber \\
&= \left( \left\{\frac{RA}{\eta_1(\vX)} - \frac{R(1 - A)}{\eta_0(\vX)}\right\}(Y - \nu_a(\vX)) + \nu_1(\vX) - \nu_0(\vX)\right)\big(\1 (\tau > \tau^*_{\kappa}) \big) \\
& \qquad \qquad + \left(\frac{R(1-A)}{\eta_0(\vX)}\right)(Y - \nu_0(\vX)) + \nu_0(\vX) \\ 
&= \xi(\vX) d^{*}_{\kappa} + \left(\frac{R(1-A)}{\eta_0(\vX)}\right)(Y - \nu_0(\vX)) + \nu_0(\vX),
\end{align*}
where $\xi(\vX) \equiv \left\{\frac{RA}{\eta_1(\vX)} - \frac{R(1 - A)}{\eta_0(\vX)}\right\}(Y - \nu_a(\vX)) + \nu_1(\vX) - \nu_0(\vX)$. 

Dropping arguments for notational convenience, we decompose the difference
\begin{align}
\estMCCARhat(\widehat d_{\kappa}^{*}) - V(d^*_{\kappa}) &= \Pn\{\ifMCCARhat(\widehat d_{\kappa}^{*})\} - \mathbb E \{Y(d^*_{\kappa})\} \nonumber \\ 
&= \Pn(\widehat \xi\widehat d^{*}_{\kappa}) - \mathbb P (\xi d_{\kappa}^{*})  \label{eq:mccarbiasestpolicy} \\
&\quad + \Pn\Big(\Big\{\frac{R(1-A)}{\widehat \eta_0}\Big\}(Y - \widehat \nu_0) + \widehat \nu_0\Big) - \mathbb P(v_0), \label{eq:mccarstandterm}
\end{align}
where (\ref{eq:mccarbiasestpolicy}) follows by iterated expectation.
(\ref{eq:mccarstandterm}) can be controlled following standard arguments. Similar to \citet{luedtke_optimal_2016}, we may expand (\ref{eq:mccarbiasestpolicy}) as
\begin{align*}
\Pn(\widehat \xi\widehat d^{*}_{\kappa}) - \mathbb P (\xi d_{\kappa}^{*}) 
&= \Pn(\widehat \xi\widehat d^{*}_{\kappa} - \tau^*_{\kappa}(\widehat d^*_{\kappa} - \kappa)) - \mathbb P (\xi d_{\kappa}^{*} - \tau^*_{\kappa}(d^*_{\kappa} - \kappa)) \\ 
&\quad + \Pn(\tau^*_{\kappa}(\widehat d^*_{\kappa} - \kappa)) - \mathbb P(\tau^*_{\kappa}(d^{*}_{\kappa} - \kappa)).
\end{align*}

By the definition of $d^{*}_{\kappa}$, $\mathbb P(d^{*}_{\kappa}) = \kappa$, rendering the last population term zero. 
To simplify the handling of the third term, we assume that there are no ties in the estimated scores $\widehat \tau(\vX)$ at the boundary $\widehat \tau^{*}_{\kappa}$, and so $\Pn(\widehat d^{*}_{\kappa}) = \kappa$ holds exactly by construction thus eliminating the third term. 
If this condition is violated, $\widehat d^{*}_{\kappa}$ may instead be implemented as a stochastic policy that randomly treats a subset of tied individuals. 
We refer to \citet{luedtke_optimal_2016} for a further discussion on how to modify the proof in this case.

We then further decompose (\ref{eq:mccarbiasestpolicy}) as
\begin{align}
\Pn(\widehat \xi \widehat d^{*}_{\kappa} - \tau^{*}_{\kappa}(\widehat d^{*}_{\kappa} - \kappa)) - \mathbb P (\xi d^{*}_{\kappa} -\tau^{*}_{\kappa}(d^*_{\kappa} - \kappa)) 
&= (\Pn - \mathbb P)((\widehat \xi  \widehat d^{*}_{\kappa} - \xi  d^{*}_{\kappa}) - \tau^{*}_{\kappa}(\widehat d^{*}_{\kappa} - d^{*}_{\kappa})) \label{eq:empiricalprocessterm} \\
&\quad + (\Pn - \mathbb P)((\xi  - \tau^{*}_{\kappa})d_{\kappa}^{*})  \label{eq:cltterm} \\ 
& \quad + \mathbb P((\widehat \xi - \tau^{*}_{\kappa}) \widehat d^{*}_{\kappa} - (\xi - \tau^{*}_{\kappa})d^{*}_{\kappa}). \label{eq:biasterm}
\end{align}

The empirical process term (\ref{eq:empiricalprocessterm}) can be handled by standard sample splitting and Donsker class arguments.
The second term (\ref{eq:cltterm}) will converge by the Central Limit Theorem. 
We now focus on the final term of the decomposition (\ref{eq:biasterm}), which may be rewritten as
\begin{align*}
\mathbb P((\widehat \xi - \tau^{*}_{\kappa}) \widehat d^{*}_{\kappa} - (\xi - \tau^{*}_{\kappa})d^{*}_{\kappa}) = \mathbb P((\widehat \xi - \xi)\widehat d^{*}_{\kappa}) + \mathbb P((\xi - \tau^{*}_{\kappa})(\widehat d^{*}_{\kappa} - d^{*}_{\kappa})).
\end{align*}

After applying iterated expectation, the first term will be controlled by exactly the same arguments as the derivation of the remainder term (\ref{app:eqmccarremainder}) from the proof of Proposition ~\ref{thm:MCCARconvergence}, resulting in the second-order term
\begin{align*}
\mathbb P((\widehat \xi - \xi)\widehat d^{*}_{\kappa})  = \mathbb E \left[ \mathbb E[\widehat \xi(\vX) - \xi(\vX)  \mid \vX] \widehat d^{*} _{\kappa}(\vX)\right] \lesssim \sum_{a \in \{0, 1\}} \| \eta_a(\vx) - \widehat \eta_a(\vx)\|\|\nu_a(\vx) - \widehat \nu_a(\vx)\|.
\end{align*}
Now, focusing on the second term,
\begin{align*}
\mathbb P((\xi - \tau_{\kappa}^{*})(\widehat d^{*}_{\kappa} - d^{*}_{\kappa})) &= \int (\tau(\vx) - \tau_{\kappa}^{*})(\widehat d^{*}_{\kappa} - d^{*}_{\kappa}) d \mathbb P(\vx) \\ 
& \leq \int | \tau (\vx) - \tau^{*}_{\kappa}| | \widehat d^{*}_{\kappa} - d^{*}_{\kappa}| d \mathbb P(\vx) \\ 
& \leq \int | \tau(\vx) - \tau_{\kappa}^{*}||\1(|\tau(\vx) -\tau^*_{\kappa}| \leq | (\widehat \tau(\vx) - \tau(\vx))  - (\widehat \tau^*_{\kappa} - \tau^*_{\kappa})|) d\mathbb P(\vx)
\\ 
& \leq \int | (\widehat \tau(\vx) - \tau(\vx))  - (\widehat \tau^*_{\kappa} - \tau^*_{\kappa})||\1(|\tau(\vx) -\tau^*_{\kappa}| \leq | (\widehat \tau(\vx) - \tau(\vx))  - (\widehat \tau^*_{\kappa} - \tau^*_{\kappa})|) d\mathbb P(\vx) \\ 
& \leq \|  (\widehat \tau(\vx) - \tau(\vx))  - (\widehat \tau^*_{\kappa} - \tau^*_{\kappa}) \|_{\infty} \mathbb P(|\tau(\vx) -\tau^*_{\kappa}| \leq \| (\widehat \tau(\vx) - \tau(\vx))  - (\widehat \tau^*_{\kappa} - \tau^*_{\kappa})\|_{\infty}) \\ 
& \lesssim \|  (\widehat \tau(\vx) - \tau(\vx))  - (\widehat \tau^*_{\kappa} - \tau^*_{\kappa}) \|_{\infty}^{1 + \alpha},
\end{align*}
where the first line holds by iterated expectation, the last inequality by Assumption ~\ref{ass:margincondition}, and the second inequality by the fact that 
\begin{align*}
|\1\{\widehat \tau(\vx) > \widehat \tau^*_{\kappa}\} - \1\{ \tau(\vx) > \tau^*_{\kappa}\} | \leq \1(|\tau(\vx) -\tau^*_{\kappa}| \leq | (\widehat \tau(\vx) - \tau(\vx))  - (\widehat \tau^*_{\kappa} - \tau^*_{\kappa})|).
\end{align*}

\clearpage
\newpage

\section{Efficiency Simulations}
\label{app:efficiencysims}

For our efficiency simulations, we use the following base settings,
\begin{align*}
X &\sim \text{Uniform}(-1, 1) \\
\pi(X) &= .4\sin(2\pi X) + .5\\ 
R \mid X &\sim \text{Bernoulli}(\pi(X)) \\
\gamma(X)  &= .4\sin(2\pi X - .72\pi) + .5\\
\lambda(X, Y) &= \gamma(X) \\ 
A \mid X &\sim \text{Bernoulli}(\gamma(X)) \\
Y \mid A &\sim \text{Bernoulli}(0.5).
\end{align*}
We do not use a treatment effect in order to have a constant residual term. Hence, $\lambda(X, Y) = P(A|X,Y) = P(A|X) = \gamma(X)$. All default constants were chosen so $\pi, \gamma, \lambda \approx 0.5$. We run 100 iterations for each configuration and report the average. For each iteration, we sample $n = 10,000$ observations. In total, all three experiments complete in 1-2 minutes on a personal laptop. All code to reproduce these experiments is available in the supplementary materials.

\paragraph{Experiment 1: Increasing $\lambda$} Let $l^* \in \{0.1, 0.15, \dots, 0.8, 0.85\}$ be the desired level for $\lambda$. We then set $\lambda(X, Y) = .09\sin(2\pi X - .72\pi) + l^*$ while keeping all other variables constant.

\paragraph{Experiment 2: Increasing $1-\pi$} Let $p^* \in \{0.1, 0.15, \dots, 0.8, 0.85\}$ be the desired level for $\pi$. We then set $\pi(X, Y) = .09\sin(2\pi X) + p^*$ while keeping all other variables constant.

\paragraph{Experiment 3: Increasing correlation between $1-\pi$ and $\lambda$} Let $c^* \in  \{0.0, 0.05, \dots, 0.9, 0.95\}$ be the correlation. We then set $\lambda(X, Y) = .4\sin(2\pi X - c^*\pi) + .5$ while keeping all other variables constant.

\clearpage
\newpage

\section{CATE Estimation Algorithms}
\label{app:sec:algos}

\subsection{CATE Estimation Algorithm for $\tmccarest$}

\begin{algorithm}
\caption{DR-MCCAR}
\begin{algorithmic}[1]
\label{alg:tmccarest}
\REQUIRE Data $\mathcal Z = \{(\vX_i, A_i, Y_i, R_i)\}_{i=1}^n$
\STATE Randomly split data into $2$ equal folds $\mathcal{Z}_1, \mathcal{Z}_2$
\FOR{$k = 1$ to $2$}
    \STATE Use $\mathcal{Z}_{-k}$ to estimate nuisance functions: $\widehat \eta_1$, $\widehat \eta_0$, $\widehat \nu_{1}$, $\widehat \nu_{0}$
    \FOR{$i \in \mathcal{Z}_k$}
        \STATE Calculate pseudo-outcome $\widetilde \phi_{\textnormal{MCCAR}}(\vX_i)$ using $\widehat \eta_1(\vX_i)$, $\widehat \eta_0(\vX_{i})$, $\widehat \nu_{1}(\vX_i)$, $\widehat \nu_{0}(\vX_i)$
    \ENDFOR
    \STATE Estimate CATE model $\widehat{\tau}_k(\vX_{i})$ predicting $\widetilde \phi_{\textnormal{MCCAR}}(\vX_i)$ from $\vX_i$ for $i \in \mathcal{Z}_k$
\ENDFOR
\STATE \textbf{return} Final CATE estimate $\widehat{\tau}(\vX) = \frac{1}{2}\sum_{k=1}^2 \widehat{\tau}_k(\vX)$
\end{algorithmic}
\end{algorithm}

\subsection{CATE Estimation Algorithms for $\tmarest$}

Recall that $\widehat \beta_{d}(\vX) = \E[Y\widehat \lambda_{d}(\vX, Y) \mid \vX] = \int y \widehat \lambda_{d}(\vX, y) d \mathbb P(Y \leq y \mid X)$ and $\widehat \gamma_{d} = \E[\widehat \lambda_{d}(\vX, Y) \mid \vX] = \int \widehat \lambda_{d}(\vX, y) d \mathbb P(Y \leq y \mid X)$. 
We may thus estimate $\widehat \beta_{d}$ and $\widehat \gamma_{d}$ through either regression or conditional density estimation.
Due to the difficulties of the latter, we propose estimating the density only for discrete outcomes and using regression and a three-way sample splitting scheme when $Y$ is continuous.
Algorithm \ref{alg:tmarestbinary} handles this first case for binary $Y$, but can be easily modified to accommodate categorical outcomes. 
Below, we let $\mu(\vX) = P(Y = y \mid \vX)$ be the conditional density.

\begin{algorithm}
\caption{DR-MAR for Binary Outcomes}
\begin{algorithmic}[1]
\label{alg:tmarestbinary}
\REQUIRE Data $\mathcal Z = \{(\vX_i, A_i, Y_i, R_i)\}_{i=1}^n$, where $Y_i \in \{0,1\}$
\STATE Randomly split data into $2$ equal folds $\mathcal{Z}_1, \mathcal{Z}_2$
\FOR{$k = 1$ to $2$}
    \STATE Use $\mathcal{Z}_{-k}$ to estimate nuisance functions: $\widehat \lambda_1$, $\widehat \mu$, $\widehat \pi$
    \FOR{$i \in \mathcal{Z}_k$}
        \STATE Set $\widehat \lambda_0(\vX_{i}, Y_i) = 1 - \widehat \lambda_1(\vX_{i}, Y_i)$, $\widehat \lambda_0(\vX_{i}, 1) = 1 - \widehat \lambda_1(\vX_{i}, 1)$
        \STATE Set $\widehat \beta_{1}(\vX_i) = \widehat \lambda_{1}(\vX_{i}, 1)\widehat \mu(\vX_{i})$, $\widehat \beta_{0}(\vX_i) = \widehat \lambda_{0}(\vX_i, 1)\widehat \mu(\vX_{i})$
        \STATE Set $\widehat \gamma_{1}(\vX_i) = \widehat \lambda_{1}(\vX_{i}, 1)\widehat \mu(\vX_{i}) +  \widehat \lambda_{1}(\vX_{i}, 0)(1-\widehat \mu(\vX_{i}))$
        \STATE Set $\gamma_{0}(\vX_i) = 1 - \gamma_{1}(\vX_i)$
        \STATE Calculate pseudo-outcome $\widetilde \varphi_{\textnormal{MAR}}(\vX_i)$ using $\widehat \lambda_1(\vX_i, Y_i)$, $\widehat \lambda_0(\vX_i, Y_i)$, $\widehat \pi(\vX_i, Y_i)$, $\widehat \beta_{1}(\vX_i)$, $\widehat \beta_{0}(\vX_i)$,  $\gamma_{1}(\vX_i)$, $\gamma_{0}(\vX_i)$
    \ENDFOR
    \STATE Estimate CATE model $\widehat{\tau}_k(\vX_{i})$ predicting $\widetilde \varphi_{\textnormal{MAR}}(\vX_i)$ from $\vX_i$ for $i \in \mathcal{Z}_k$
\ENDFOR
\STATE \textbf{return} Final CATE estimate $\widehat{\tau}(\vX) = \frac{1}{2}\sum_{k=1}^2 \widehat{\tau}_k(\vX)$
\end{algorithmic}
\end{algorithm}

Algorithm \ref{alg:tmarestbinary} is not valid for continuous outcomes. 
To avoid conditional density estimation, we instead estimate $\widehat \beta$ and $\widehat \gamma$ using a nested cross-fitting procedure. 
There are many ways to split the data this way.
For example, one could split the data into thirds and estimate six different CATE models (one for each permutation of which fold estimates nuisances, which estimates $\widehat \beta, \widehat \gamma$ and which estimates the CATE).
To reduce computational complexity, Algorithm \ref{alg:drmarcontoutcomes} maintains two main splits for two different CATE models $\widehat \tau_{1}, \widehat \tau_{2}$, but further splits each nuisance fold in half to separately estimate $\widehat \lambda$ versus $\widehat \beta$, $\widehat \gamma$ in nested cross-fitting loop.  

\begin{algorithm}
\caption{DR-MAR for Continuous Outcomes}
\begin{algorithmic}[1]
\label{alg:drmarcontoutcomes}
\REQUIRE Data $\mathcal Z = \{(\vX_i, A_i, Y_i, R_i)\}_{i=1}^n$, where $Y_i \in \mathbb{R}$
\STATE Randomly split data into $2$ equal folds $\mathcal{Z}_1, \mathcal{Z}_2$
\FOR{$k = 1$ to $2$}
    \STATE Use  $\mathcal{Z}_{-k}$ to estimate $\widehat \pi$
    \STATE Split $\mathcal{Z}_{-k}$ into two equal halves: $\mathcal{Z}_{-k}^{(1)}$, $\mathcal{Z}_{-k}^{(2)}$
    \FOR{$j = 1$ to $2$}
        \STATE Use $\mathcal{Z}_{-k}^{(-j)}$ to estimate nuisance function $\widehat{\lambda}_1^{(j)}$
        \FOR{$i \in \mathcal{Z}_{-k}^{(j)}$}
            \STATE Estimate $\widehat{\beta}_1^{(j)}(\vX_i)$, $\widehat{\beta}_0^{(j)}(\vX_i)$, $\widehat{\gamma}_1^{(j)}(\vX_i)$ using $\widehat{\lambda}_1^{(j)}$
        \ENDFOR
    \ENDFOR
    \FOR{$i \in \mathcal{Z}_k$}
        \STATE Set $\widehat{\beta}_1(\vX_i) = \frac{1}{2}\sum_{j=1}^2 \widehat{\beta}_1^{(j)}(\vX_i)$ and $\widehat{\beta}_0(\vX_i) = \frac{1}{2}\sum_{j=1}^2 \widehat{\beta}_0^{(j)}(\vX_i)$
        \STATE Set $\widehat{\gamma}_1(\vX_i) = \frac{1}{2}\sum_{j=1}^2 \widehat{\gamma}_1^{(j)}(\vX_i)$
        \STATE Set $\widehat{\lambda}_1(\vX_i, Y_i) = \frac{1}{2}\sum_{j=1}^2 \widehat{\lambda}_1^{(j)}(\vX_i, Y_i)$
        \STATE Set $\widehat \lambda_0(\vX_{i}, Y_i) = 1 - \widehat \lambda_1(\vX_{i}, Y_i)$
        \STATE Set $\widehat \gamma_0(\vX_i) = 1 - \widehat \gamma_{1}(\vX_i)$ 
        \STATE Calculate pseudo-outcome $\widetilde \varphi_{\textnormal{MAR}}(\vX_i)$ using $\widehat \lambda_1(\vX_i, Y_i)$, $\widehat \lambda_0(\vX_i, Y_i)$, $\widehat \pi(\vX_i, Y_i)$, $\widehat \beta_{1}(\vX_i)$, $\widehat \beta_{0}(\vX_i)$,  $\widehat \gamma_{1}(\vX_i)$, $\widehat \gamma_{0}(\vX_i)$
    \ENDFOR
    \STATE Estimate CATE model $\widehat{\tau}_k(\vX_{i})$ predicting $\widetilde \varphi_{\textnormal{MAR}}(\vX_i)$ from $\vX_i$ for $i \in \mathcal{Z}_k$
\ENDFOR
\STATE \textbf{return} Final CATE estimate $\widehat{\tau}(\vX) = \frac{1}{2}\sum_{k=1}^2 \widehat{\tau}_k(\vX)$
\end{algorithmic}
\end{algorithm}

\clearpage
\newpage
\section{Synthetic Experiment Details}
\label{app:sec:convergeneexp}

All code to reproduce the following experiments is available at \href{https://github.com/jesund/learningwithmissingtreatments}{https://github.com/jesund/learningwithmissingtreatments}.

\subsection{Convergence Experiment}

We generate $n = 1,000$ samples using the data-generating processes described in Section \ref{sec:syntheticexperiments}. 
We calibrated the constant $C$ in the missingness functions $\pi$ empirically to match target observation rates. 
We report our results with observation rates of $25\%$ and $75\%$ in the main text, and below report our results with the other observation rates we tested in Fig. \ref{fig:app:cateoned}.
For each observation rate in the MAR and MCCAR setting, we ran 500 iterations. 
All nuisance functions used their true values and were not estimated. 
We then added randomly sampled noise $\epsilon \sim N(-n^{-\alpha}, n^{-2\alpha})$ to each nuisance function, where $\alpha$ is the rate of convergence, i.e. setting $\alpha = \frac{1}{2}$ corresponds to $\sqrt n-$rates of convergence.
The noisy nuisance functions were then used to compute pseudo-outcomes $\widetilde \varphi_{\textnormal{MAR}}$ and $\widetilde \varphi_{\textnormal{MCCAR}}$.
Pseudo-outcomes for oracle models were computed using the noise-free nuisance functions.
We used a \verb|smooth.spline| from base R with the default parameters to estimate the CATE for all models, including the oracle, by regressing the pseudo-outcomes on $X$.
We then calculated the RMSE on a test set of size $1,000$ and report the median over the 500 iterations.
All experiments were run on a personal laptop with 12 CPU cores and took 5-10 minutes to complete.

\begin{figure}
    \centering
    \begin{subfigure}[t]{0.9\textwidth}
        \centering
        \includegraphics[width=\textwidth]{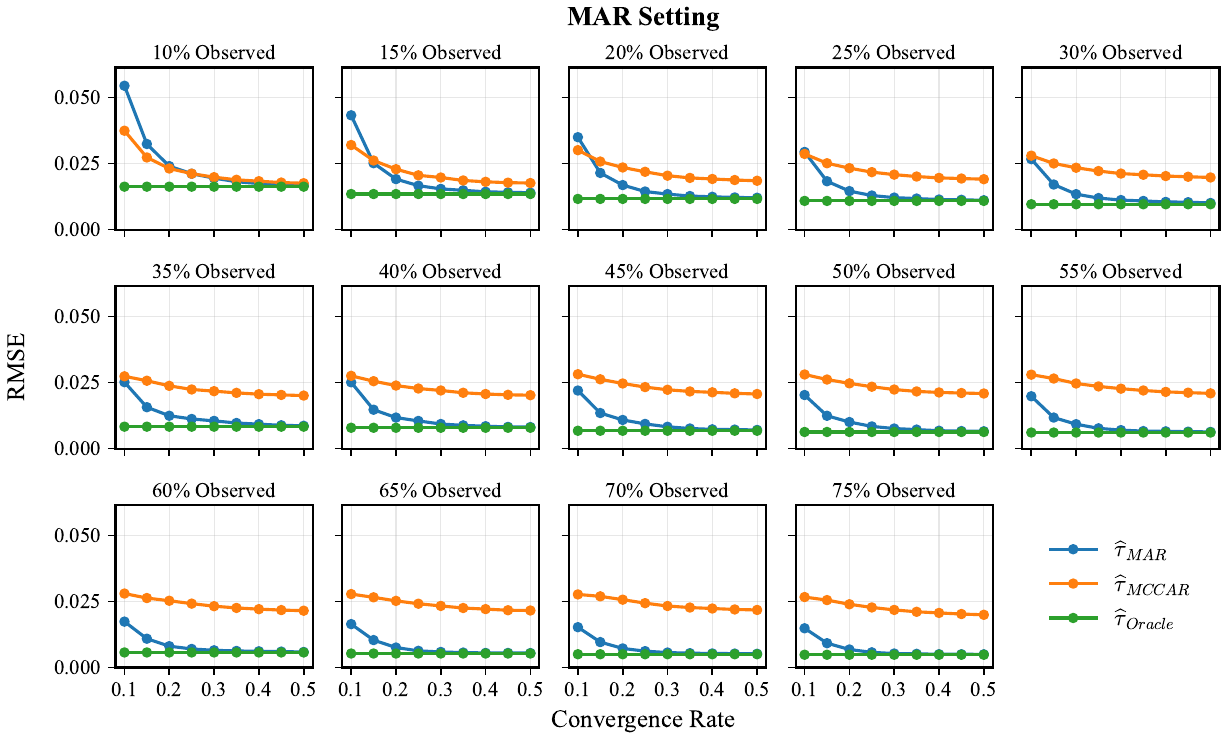}
        \label{fig:cateoned_mar}
    \end{subfigure}
    
    \vspace{0.01cm}
    
    \begin{subfigure}[t]{0.9\textwidth}
        \centering
        \includegraphics[width=\textwidth]{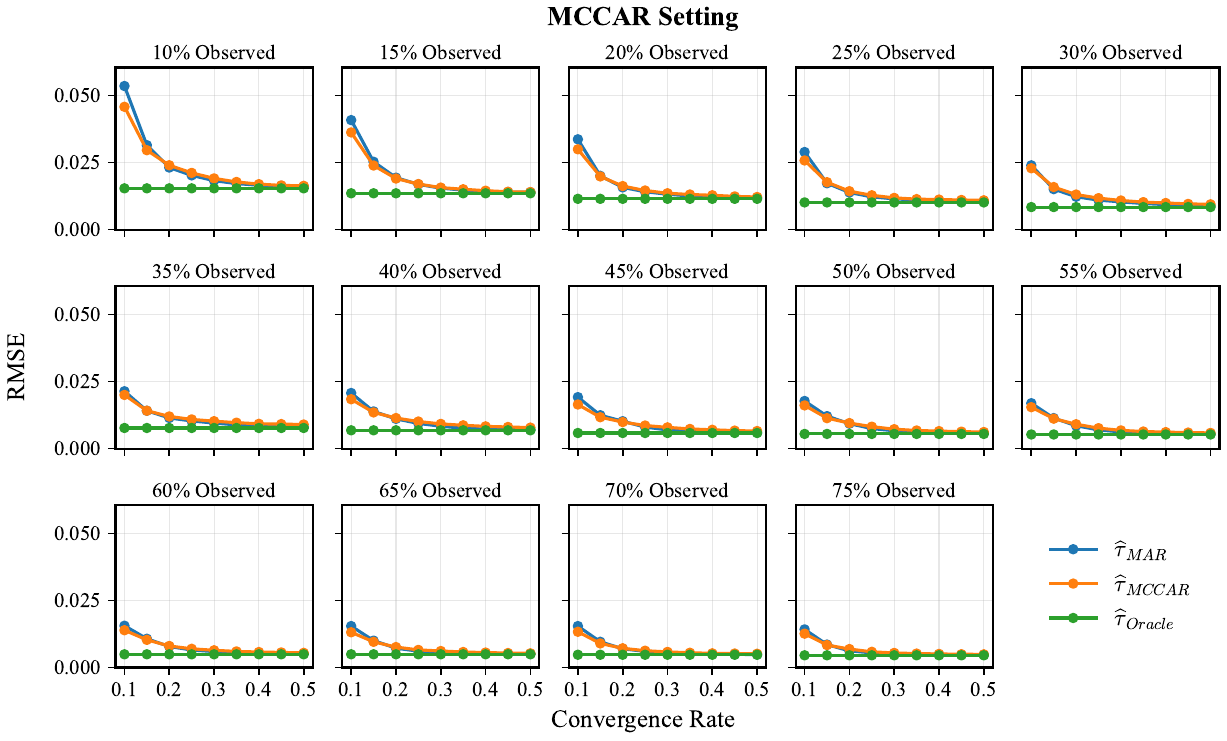}
        \label{fig:cateoned_mccar}
    \end{subfigure}
    
    \caption{Median RMSE of CATE estimates comparing $\tmarest$ (blue), $\tmccarest$ (orange), and $\toracle$ (green) in the MAR (a) and MCCAR (b) settings over 500 iterations against the rate of noise added to all nuisance functions. In the MAR setting (top), $\tmccarest$ remains biased even after more treatments are observed, while in the MCCAR setting (bottom) both estimators are unbiased and $\tmarest$ achieves slightly faster convergence. This pattern holds across all observation rates including $25\%$ and $75\%$.}
    \label{fig:app:cateoned}
\end{figure}

\subsection{High-Dimensional Experiment}

Our two high-dimensional experiments generate data following the MAR and MCCAR data-generating processes given in Section \ref{sec:highdimexp}. 
We run 50 iterations for each sample size $n \in \{1,000, 2,000, 5,000, 10,000\}$ 
under both settings.
For each iteration, we estimate the CATE following Algorithm \ref{alg:tmarestbinary}. 
We use \verb|RandomForest| for all nuisance function estimation.
All propensity scores are clipped to be in $[0.01, 0.99]$.
We found that using a shallower \verb|RandomForest| for the CATE model worked best based on preliminary experiments, and so we set \verb|max_depth = 3|.
We compare our estimates to an oracle CATE that uses the true values for all nuisance functions in the input to the \verb|RandomForest|.
All pseudo-outcomes are clipped to be within the 1st and 99th quantiles.
We generate a separate test sample of the same size for each iteration. 
We report the average RMSE over each iteration for each sample size and setting.
All experiments were run on a personal laptop with 12 CPU cores and took 10-20 minutes to complete.

\subsubsection{Misspecification Experiment}
To test the robustness properties of our proposed estimators, we run a variant of the high-dimensional experiment.
To introduce model misspecification, we systematically train nuisance functions with $\verb|RandomForest|$ with $\verb|max_depth=1|$.
The shallow trees are unable to capture the complexity of the data-generating process in Section~\ref{sec:highdimexp}, as each nuisance includes multiple variables and some have interaction terms.
All other hyperparameters and experimental details remain the same as described above.

\paragraph{MCCAR Misspecification Results} Table~\ref{tab:misspechighdimmccar} shows the results of the experiment in the MCCAR setting. Following Proposition~\ref{thm:MCCARconvergence}, if $\nu_d$ or $\eta_d$ is misspecified, the MCCAR estimator still converges. On the other hand, the MCCAR estimator is not robust to misspecification of both $\nu_d$ and $\eta_d$. 

\begin{table}[ht]
    \centering
    \begin{tabular}{ccccc}
        \toprule
        $N$ & $\widehat \tau_{MCCAR}$ & $\widetilde \nu_d$ & $\widetilde \eta_d$ & $\widetilde \nu_d, \widetilde \eta_d$  \\
        \midrule
        1000  & 0.095 & 0.078 & 0.155 & 0.130 \\
        5000  & 0.021 & 0.019 & 0.040 & 0.048 \\
        10000 & 0.014 & 0.014 & 0.031 & 0.044 \\
        \bottomrule
    \end{tabular}
    \caption{Average CATE RMSE of the MCCAR estimator under misspecification, where $\tilde f$ denotes a misspecified $f$. Under Proposition~\ref{thm:MCCARconvergence}, the model is robust to misspecification in $\eta_d$ or $\nu_d$, but not both simultaneously.}
    \label{tab:misspechighdimmccar}
\end{table}

\section{Semi-Synthetic Experiment Details}
\label{app:sec:semisynthetic}

All instructions for downloading the data and code to run the semi-synthetic experiments are provided at \href{https://github.com/jesund/learningwithmissingtreatments}{https://github.com/jesund/learningwithmissingtreatments}.
All experiments were run on a computing cluster. 
MTRNET was tuned and trained on an NVIDIA L40S GPU, while other models were trained using nodes with 24 CPUs. For a single replication, hyperparameter tuning and training takes 1-2 hours for MTRNET versus 5-10 minutes for the other methods.
\subsection{Implementation Details}

\subsubsection{Data preprocessing}

\paragraph{Voting} This dataset comes from a 2008 field experiment studying the effects of social pressure on voter turnout \citep{gerber_social_2008}. We focused on two experimental arms: the ``Neighbors'' treatment ($A = 1$), where potential voters received mailings showing their neighbors' voting records, and the control group ($A = 0$), which received no mailings. The outcome $Y$ is a binary indicator for whether the individual voted in the August 2006 primary election. We preprocessed the data by converting the year of birth to age, removing the household ID and cluster ID, and converting categorical variables to dummy variables using one-hot encoding, dropping the first category of each variable to avoid multicollinearity.

\paragraph{Tennessee's Student Teacher Achievement Ratio (STAR) Project} 
This dataset comes from a four-year longitudinal study examining the effect of class size on student achievement \citep{achilles_tennessees_2008}. We focused on kindergarten students, comparing two experimental conditions: small class (13-17 students) and regular class (22-25 students), excluding the regular-with-aide condition to maintain binary treatment. We preprocessed the data by filtering for kindergarten students only and excluded students in the regular-with-aide class type. We created a binary treatment indicator where regular class was coded as 0 and small class as 1. We removed students with missing test scores and created a composite outcome variable $Y$ by summing the kindergarten test scores across four domains: reading, math, listening, and word skills.

We retained key covariates including gender, race, birth month, school, teacher demographics (gender, race, years of experience, education level, career ladder), and student characteristics (attendance, free lunch eligibility, grade repetition, special education status). Categorical variables were converted to dummy variables using one-hot encoding, and missing values for attendance variables were imputed with the median values. 

\paragraph{Generating Treatment Missingness} We artificially induce treatment missingness by generating observation scores $s_i$  for each individual $i$ and then sampling binary indicators $R_i$ based on these scores. 
We base the missingness mechanism on the most predictive covariate $X^\star$ for each dataset, identified by feature importance from a \verb|RandomForest|: years of teaching experience (STAR) and age (Voting).

\subparagraph{MAR}
For the continuous outcome dataset (STAR), we generate observation scores using both outcomes and $X^\star$ for each individual $i$:
\begin{align*}
q &= F_{Y}^{-1}(r) \\
s_i &= \1(Y_i < q) (X_{i}^{\star} + 1) + 0.1
\end{align*}
where $F_Y^{-1}(r)$ is the $r$-th quantile of the outcome distribution, $r$ is the observation rate, and we add one to $X_{i}^{\star}$ to set the minimum above zero. For the binary outcome dataset (Voting), we use:
\begin{align*}
s_i &= r Y_i (X_{i}^{\star} - \min(X^{\star}) + 1) + 0.1
\end{align*}
where we transform $X^{\star}$ so that the minimum value is one, as in the continuous setting.
The baseline probability of 0.1 under both mechanisms ensures that all treatment assignments have some chance of being observed. 

\subparagraph{MCCAR}
We generate missingness dependent only on $X^{\star}$ by setting observation scores based on thresholds of $X^{\star}$:
\begin{align*}
q &= F_{X^{\star}}^{-1}(r) \\
s_i &= 0.8\1(X_{i}^{\star} < q) + 0.1
\end{align*}
where $F_{X^\star}^{-1}(r)$ is the $r$-th quantile of $X^{\star}$ and $r$ is again the observation rate. 

\paragraph{Sampling} 
After computing observation propensity scores $s_i$ for each individual, we employ an incremental sampling algorithm to select which treatment assignments are observed. 
This ensures that as the observation rate $r$ increases, the set of observed treatments expands monotonically rather than changing completely. 
We first normalize scores to sampling probabilities: $p_i = \frac{s_i}{\sum_{j=1}^{n} s_j}$.  We then sample a random permutation of all $n$ indices without replacement using \texttt{numpy.random.choice} with probabilities $p_i$, producing an ordered sequence  representing the sampling priority. 
To achieve an observation rate of $r$ (e.g., $r=0.3$ for 30\% observed), we set $R_i = 1$ for the first $\lfloor nr \rfloor$ observations in the sequence and $R_i = 0$ for the remainder.
This nested structure ensures that increasing $r$ adds new observations while retaining all previously observed ones.

\subsubsection{Modeling}

We compare the results between the following models in the main text:
\begin{enumerate}[topsep=0pt,partopsep=-0.2ex,parsep=0.2ex]
    \item \textbf{DR-MAR} Algorithm \ref{alg:tmarestbinary} for the Voting 
          dataset and Algorithm \ref{alg:drmarcontoutcomes} for the STAR dataset.
    \item \textbf{DR-MCCAR} Algorithm \ref{alg:tmccarest}.
    \item \textbf{MTRNET} Official implementation \citet{kuzmanovic_estimating_2023} (https://github.com/mkuzma96/MTRNet).
    \item \textbf{OR} Outcome regression models fit on the nuisance functions using Eq.~\ref{eq:cccate_estimand} in the MCCAR setting and Eq.~\ref{eq:marcateestimator} in the MAR setting. We do not use a second-stage regression.
    \item \textbf{PW-Learner} which computes pseudo-scores using only propensity scores \citep{curth_nonparametric_2021} and then employs a second stage regression model to estimate the CATE. Below we give the full estimators we use in each setting.
\end{enumerate}
And include the following additional models for comparison in this Appendix:
\begin{enumerate}
    \setcounter{enumi}{5}
    \item \textbf{DR-CC} Complete-case doubly robust learner using only observations with $R_i = 1$.
    \item \textbf{Risk Model} Outcome regression ignoring treatment, predicting $\mathbb{E}[Y|X]$.
\end{enumerate}

\subsubsection{PW-Learner}

\paragraph{MAR Setting} In the MAR setting, we compute the following pseudo-score:
\begin{align*}
\frac{AYR}{\widehat \gamma_1(\mathbf X)\widehat \pi(\mathbf X,Y)} - \frac{(1-A)YR}{\widehat \gamma_0(\mathbf X)\widehat \pi(\mathbf X,Y)},
\end{align*}
where $\widehat \gamma_1$ is estimated by first estimating $\widehat \lambda_1$, and then regression $\widehat \lambda_1$ on $\vx$ (as is done for the DR-MAR estimator).  
We use the same sample splitting scheme as the DR-MAR estimator.
 Under MAR Assumptions~\ref{assum:mar} for a specific $\vx$, it follows that
\begin{align*}
\mathbb E \left[\frac{AYR}{\gamma_1(\vx)\pi(\vx, Y)} \mid \mathbf X = x\right] = \mathbb E \left[\frac{Y\mathbb E \left[AR \mid \vX  = \vx, Y\right]}{\gamma_1(\vx)\pi(\vx, Y)} \mid \mathbf X = \vx \right] = \frac{\mathbb E \left[Y \lambda_1(\vx, Y)\mid \vX = \vx\right]}{\gamma_1(\vx)} = \frac{\beta_1(\vx)}{\gamma_{1}(\vx)},
\end{align*}
which is exactly the first term of the identified cate in Eq.~\ref{eq:marcateestimator}. The second term follows a similar proof.

\paragraph{MCCAR Setting} In the MCCAR setting under Assumption~\ref{assum:mccar}, we use the following pseudo-score:
\begin{align*}
\frac{AYR}{\widehat \eta_1(\vx)} - \frac{(1-A)YR}{\widehat \eta_0(\vx)}.
\end{align*}
In practice, $\widehat \eta_d$ may be estimated using separate models for $P(A  = d\mid \vX = \vx, R = 1)$ and $P(R = 1 \mid \vX = \vx)$, which we find performs slightly better than estimating the joint probability of both treatment and missingness. 
We follow the same sample splitting scheme as the DR-MCCAR estimator. It follows that
\begin{align*}
\mathbb E \left[\frac{AYR}{\eta_1(\vx)} \mid \vX = \vx \right] = \frac{\mathbb E \left[Y \mid A = 1, \vX = \vx\right]\eta_1(\vx)}{\eta_1(\vx)} = \mathbb E \left[Y \mid A = 1, \vX = \vx \right],
\end{align*} 
where we used the fact that $\eta_1(\vx) = P(R = 1, A = 1 \mid \vX = \vx)$. 
The second term follows a similar proof.

\subsubsection{Hyperparameter tuning}
\label{app:sec:semisyntuning}

\paragraph{Outcome Regressions, PW-Learner, and DR-Learner Variants and Risk Models}

We use an ensemble for all nuisance function estimation and pseudo-outcome regression of five base learners: ridge/logistic regression, elastic net/lasso, random forest, XGBoost, and K-nearest neighbors.
Propensity scores for all PW-Learner and DR-Learner variants are clipped to [0.1, 0.9].
After computing pseudo-outcomes, we winsorize at the 5th and 95th percentiles for the STAR dataset to reduce the influence of extreme values. 
For Voting, no winsorization is applied due to the larger sample size providing 
more stable estimates.

\begin{table}[h]
\centering
\caption{Hyperparameter grid for base learners}
\label{tab:hyperparameters}
\small
\begin{tabular}{lll}
\toprule
\textbf{Model} & \textbf{Hyperparameter} & \textbf{Values} \\
\midrule
\multirow{2}{*}{Ridge / Logistic} & $\alpha$ / $C$ & $\{0.001, 0.01, 0.1, 1, 10, 100\}$ \\
& Penalty & L2 (ridge penalty) \\
\midrule
\multirow{2}{*}{Elastic Net / Lasso} & $\alpha$ / $C$ & $\{0.01, 0.1, 1, 10\}$ \\
& $L_1$ ratio & $\{0.1, 0.3, 0.5, 0.7, 0.9\}$ (regression) / 1.0 (classification) \\
\midrule
\multirow{4}{*}{Random Forest} & \texttt{n\_estimators} & 500 \\
& \texttt{max\_depth} & $\{3, 5, 10, \text{None}\}^*$ \\
& \texttt{max\_features} & $\{\text{sqrt}, \text{None}\}$ \\
& \texttt{min\_samples\_split} & $\{2, 5\}^*$ \\
\midrule
\multirow{3}{*}{XGBoost} & \texttt{n\_estimators} & $\{100, 300\}$ \\
& \texttt{max\_depth} & $\{3, 5\}$ \\
& \texttt{learning\_rate} & $\{0.01, 0.1\}$ \\
\midrule
\multirow{2}{*}{K-Nearest Neighbors} & $k$ & $\{3, 5, 7, 11, 15\}$ \\
& \texttt{weights} & $\{\text{uniform}, \text{distance}\}$ \\
\bottomrule
\end{tabular}
\vspace{2mm}

\footnotesize{$^*$For Voting: \texttt{max\_depth} $\in \{3, 5, 10, 20\}$, \texttt{min\_samples\_split} $\in \{10, 20\}$}
\end{table}

We use 3-fold cross-validation for hyperparameter tuning. 
Classification models are optimized for negative log-loss, while regression models are optimized for negative mean squared error. 
We standardize all continuous features for ridge/logistic regression, elastic net/lasso, and KNN models.

Final ensemble predictions are weighted averages of the five base learner predictions.
For STAR, we use uniform ensemble weights.
For Voting, we learn optimal weights using non-negative least squares on 2-fold cross-validated predictions, normalized to sum to one.

\paragraph{MTRNET} Hyperparameters are selected using Optuna \citep{akiba2019optuna}with the Tree-structured Parzen Estimator (TPE) sampler. 
We use 100 trials for STAR and 50 trials for Voting (due to computational cost) using the search space in Table~\ref{tab:mtrnet_hyperparameters}.
We hold out 20\% of the training data as a validation set and optimize for AUPEC on the validation split.
All features are standardized to zero mean and unit variance before training.
We then retrain the tuned MTRNet on the full training set using the selected hyperparameters.

\begin{table}[h]
\centering
\caption{MTRNet hyperparameter search space}
\label{tab:mtrnet_hyperparameters}
\small
\begin{tabular}{ll}
\toprule
\textbf{Hyperparameter} & \textbf{Search Space} \\
\midrule
Representation layer size & $\{50, 100, 200\}$ \\
Hypothesis layer size & $\{50, 100, 200\}$ \\
Learning rate & $\{0.0001, 0.0005, 0.001, 0.005, 0.01\}$ \\
Dropout rate & $\{0.1, 0.2, 0.3\}$ \\
Training iterations & $\{100, 200, 300\}$ \\
Batch size & $\{100, 300, 500, 1000, 1500\}^*$ \\
$\alpha$ (IPW weight) & $\{10^{k/2} : k \in \{-4, -3.5, \ldots, 2\}\}$ \\
$\beta$ (representation penalty) & $\{10^{k/2} : k \in \{-4, -3.5, \ldots, 2\}\}$ \\
$\lambda$ (regularization) & $\{0.00005, 0.0001, 0.0005\}$ \\
\bottomrule
\end{tabular}
\vspace{2mm}

\footnotesize{$^*$For Voting: batch size $\in \{1000, 2500, 5000\}$}
\end{table}

\subsubsection{Evaluation}

We split the data 80-20 for training and testing, and evaluate all methods 
on the held-out test set.

\paragraph{Policy Value and AUPEC}
For a given treatment budget $\kappa \in [0,1]$ , we compute the policy value as follows. 
First, we treat the decision rule generated by $\widehat \tau$ as fixed because it was trained on a separate sample.
We then let $\widehat \tau^*_{\kappa} = \max\{F_{\widehat{\tau}}^{-1}(\kappa), 0\}$ be the treatment threshold. We assign treatment to individuals with $\widehat{\tau}(\mathbf{X}_i) > \tau^*_{\kappa}$, creating treatment decisions $D_i \in \{0,1\}$. 
We estimate the policy value by $V_{\kappa}(\widehat \tau)= \mathbb E \left[Y(1) \mid D = 1\right]P(D = 1) + \E[Y(0)\mid D = 0] P(D = 0)$, which we compute empirically using the observed outcomes under random treatment assignment. 
The AUPEC summarizes performance across all budgets: $\text{AUPEC}(\widehat{\tau}) = \int_0^1 \left[ V_{\kappa}(\widehat{\tau}) - V_{\kappa}^{\text{random}} \right] d\kappa$, where $V_\kappa^{\text{random}} = \kappa  \mathbb{E}[Y(1)] + (1-\kappa)  \mathbb{E}[Y(0)]$ is the expected policy value under random assignment.

\paragraph{Experimental Design}
For each dataset and missingness mechanism, we vary the observation rate $r \in \{0.1, 0.2, \ldots, 0.9\}$ to simulate different levels of treatment missingness. For each combination of dataset, mechanism, and observation rate, we run 10 replications (STAR) or 5 replications (Voting) with different random seeds. We conduct three separate robustness experiments, each varying:
\begin{itemize}
    \item Missingness generation: varying the random permutation used to sample
    \item Train-test split: varying the random seed for data splitting
    \item Model initialization: varying random seeds for model training
\end{itemize}
We report mean AUPEC for each configuration, allowing us to robustly assess both average performance and variability across different sources of randomness.

\subsection{Additional Results}
\label{app:sec:semisyntheticresults}

\subsubsection{Comparison to Risk Model and DR-CC}

\begin{figure}[H]
    \centering
    \includegraphics[width=0.5\textwidth]{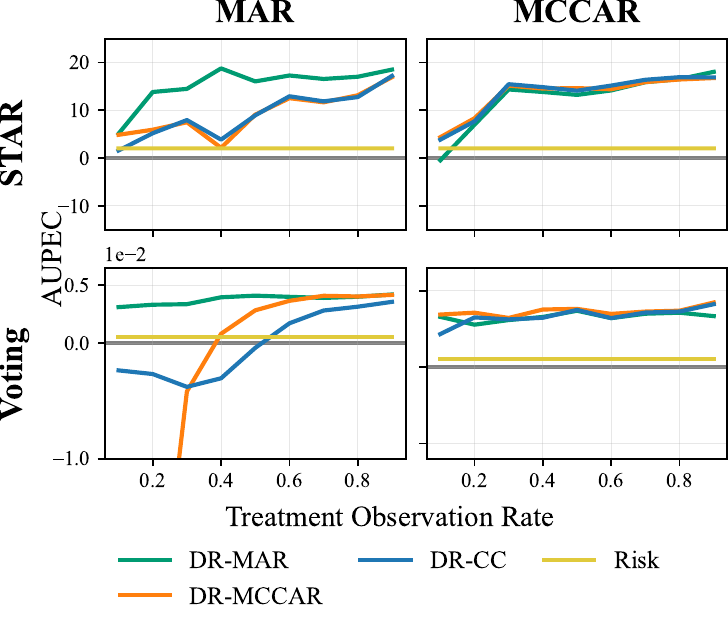}
    \caption{Average AUPEC of CATE estimators (zero is no gain over a random policy), averaged over 10 seeds (STAR) and 5 seeds (Voting) under MAR and MCCAR missingness. The DR-MAR estimator performs better in all settings. All models, except the risk model, perform well in the MCCAR setting.}
\end{figure}

We compare the results for the standard doubly-robust DR-Learner using only complete observations, and the Risk Models which predict the risk of an adverse outcome. We tested two rules based on the risk model: (1) treat those with the highest risk of the adverse outcome, and (2) treat those in the middle of the risk distribution. These rules reflect different heuristics policy-makers use in practice when assigning treatment. We compare our results to the best performing rule for each dataset.

\subsubsection{Robustness Checks for AUPEC Results}

\begin{figure}[H]
    \centering
    \begin{subfigure}[b]{0.48\textwidth}
        \centering
        \includegraphics[width=\textwidth]{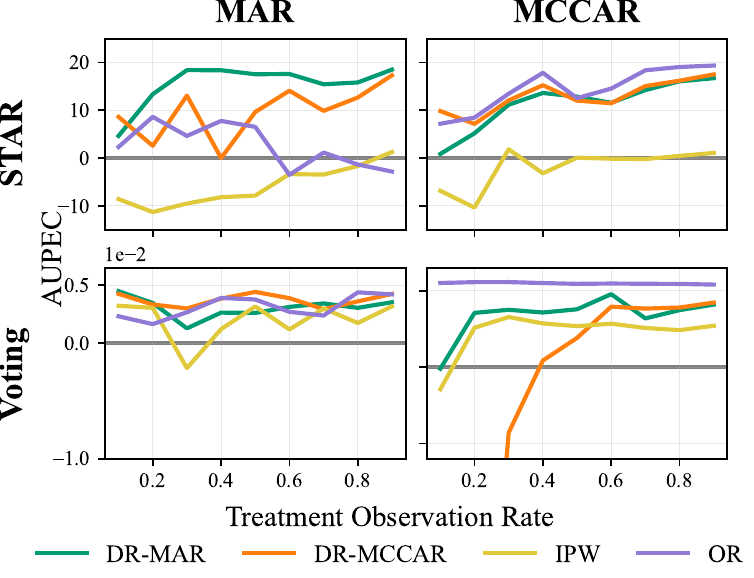}
        \caption{Changing model seed}
        \label{fig:aupec_model}
    \end{subfigure}
    \hfill
    \begin{subfigure}[b]{0.48\textwidth}
        \centering
        \includegraphics[width=\textwidth]{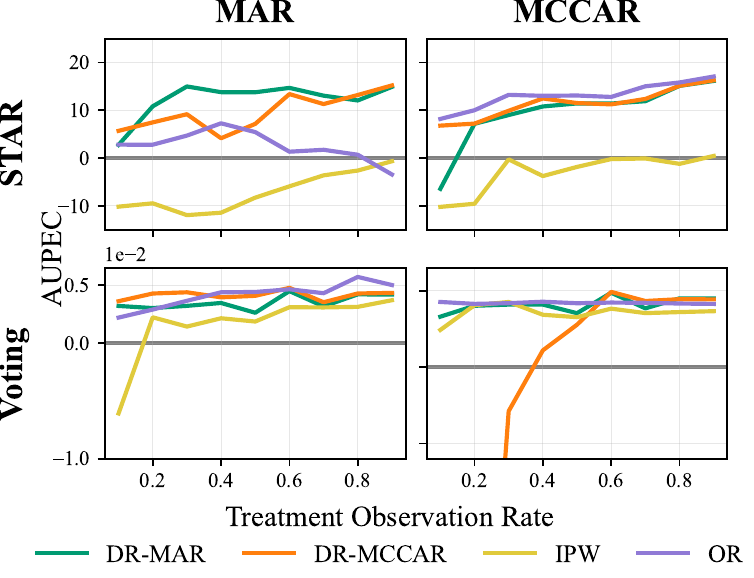}
        \caption{Changing train-test split seed}
        \label{fig:aupec_split}
    \end{subfigure}
    \caption{AUPEC with 95\% confidence intervals, where zero indicates no gain over random policy. Results show average performance across treatment observation rates for STAR (10 iterations) and Voting (5 iterations) datasets under MAR and MCCAR missingness. Left: varying model initialization seed. Right: varying train-test split seed. Train-test split variation introduces substantially more uncertainty than model seed variation. The results support our main conclusions: in the MAR setting, DR-MAR outperforms all other estimators, while in the MCCAR setting all models are competitive. We omit MTRNET from this comparison due to computational costs.}
    \label{fig:aupec_results}
\end{figure}

\clearpage
\newpage
\subsubsection{Voting - MAR}

\begin{table}[H]
    \centering
    \begin{tabular}{cccccccccc}
\toprule
 &  & DR-CC & DR-MAR & DR-MCCAR & IPW & MTRNET & OR & Random & Risk \\
Obs. Rate & $\kappa$ &  &  &  &  &  &  &  &  \\
\midrule
\multirow[t]{3}{*}{20\%} & 10\% & 0.303 & 0.306 & 0.301 & 0.306 & 0.304 & \textbf{0.307} & 0.305 & 0.305 \\
 & 25\% & 0.312 & 0.320 & 0.309 & 0.319 & 0.318 & \textbf{0.323} & 0.317 & 0.319 \\
 & 50\% & 0.332 & 0.341 & 0.309 & 0.340 & 0.340 & \textbf{0.346} & 0.337 & 0.337 \\
\cline{1-10}
\multirow[t]{3}{*}{50\%} & 10\% & 0.303 & \textbf{0.309} & 0.304 & 0.306 & 0.304 & 0.307 & 0.305 & 0.305 \\
 & 25\% & 0.315 & 0.321 & 0.318 & 0.318 & 0.317 & \textbf{0.322} & 0.317 & 0.319 \\
 & 50\% & 0.336 & 0.343 & 0.340 & 0.340 & 0.340 & \textbf{0.345} & 0.337 & 0.337 \\
\cline{1-10}
\multirow[t]{3}{*}{80\%} & 10\% & 0.306 & 0.308 & \textbf{0.309} & 0.305 & 0.305 & 0.307 & 0.305 & 0.305 \\
 & 25\% & 0.320 & 0.320 & 0.320 & 0.317 & 0.319 & \textbf{0.322} & 0.317 & 0.319 \\
 & 50\% & 0.340 & 0.343 & 0.342 & 0.340 & 0.343 & \textbf{0.345} & 0.337 & 0.337 \\
\cline{1-10}
\bottomrule
\end{tabular}

    \caption{Policy values on the test set, averaged over 5 iterations varying the missingness seed. Left columns show the percentage of treatments observed and the available treatment budget $\kappa$. Highest policy value in each row is in bold. DR-MAR outperforms all other methods at low treatment observation rates and is competitive with higher observation rates.}
\end{table}

\begin{figure}[H]
    \centering
    \includegraphics[width=0.9\textwidth]{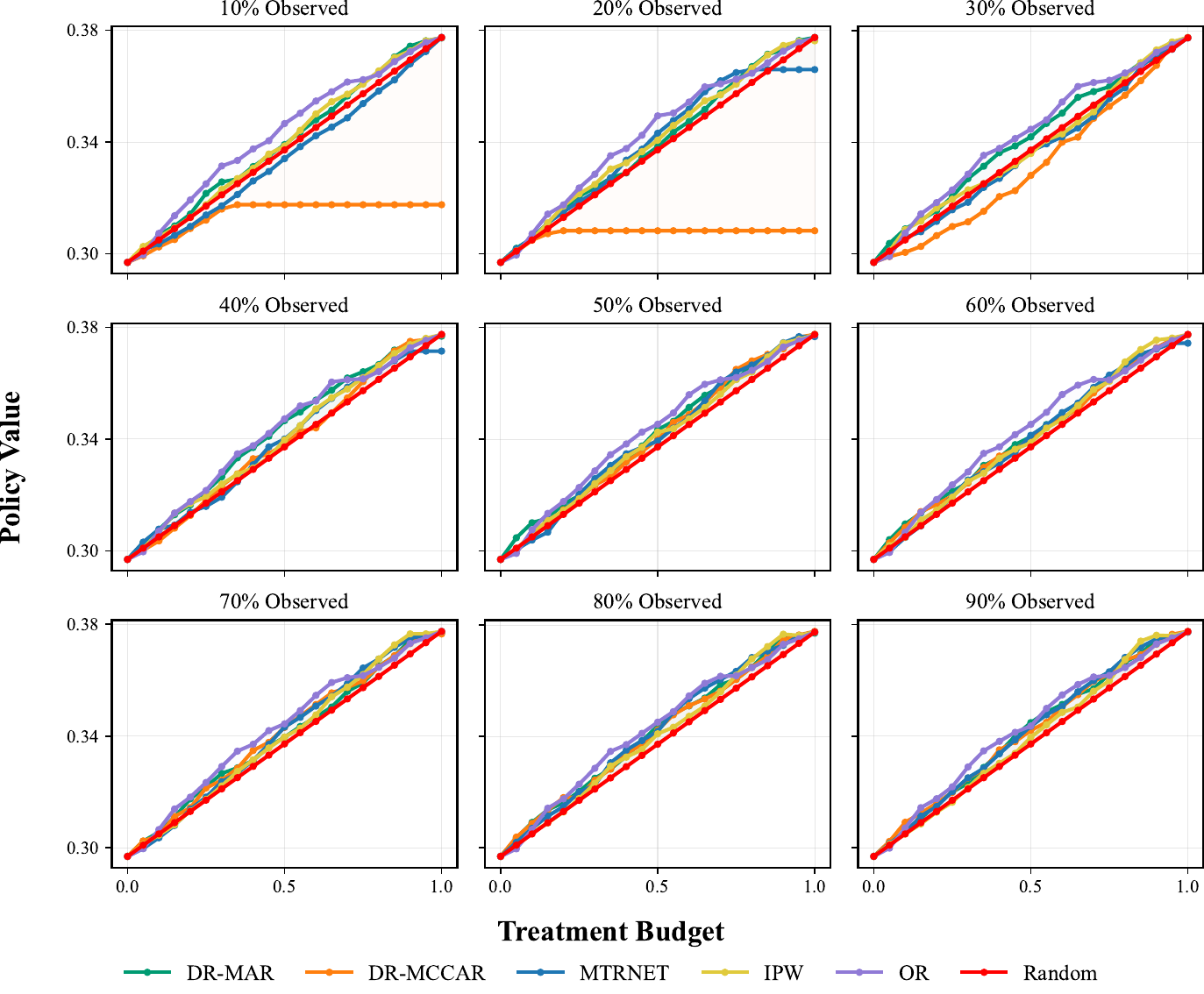}
    \caption{Example policy values across budgets and treatment observation rates. At low treatment observation rates, DR-MCCAR fails to use the entire treatment budget. DR-MAR outperforms other methods at low observation rates, with all methods converging at high observation rates}
\end{figure}

\subsubsection{Voting - MCCAR}

\begin{table}[H]
    \centering
    \begin{tabular}{cccccccccc}
\toprule
 &  & DR-CC & DR-MAR & DR-MCCAR & IPW & MTRNET & OR & Random & Risk \\
Obs. Rate & $\kappa$ &  &  &  &  &  &  &  &  \\
\midrule
\multirow[t]{3}{*}{20\%} & 10\% & \textbf{0.308} & 0.307 & 0.306 & 0.306 & 0.306 & 0.307 & 0.305 & 0.305 \\
 & 25\% & \textbf{0.321} & 0.321 & 0.320 & 0.318 & 0.318 & 0.318 & 0.317 & 0.319 \\
 & 50\% & \textbf{0.341} & 0.339 & 0.340 & 0.341 & 0.341 & 0.340 & 0.337 & 0.337 \\
\cline{1-10}
\multirow[t]{3}{*}{50\%} & 10\% & 0.308 & \textbf{0.308} & 0.306 & 0.307 & 0.306 & 0.305 & 0.305 & 0.305 \\
 & 25\% & 0.321 & \textbf{0.322} & 0.319 & 0.319 & 0.319 & 0.318 & 0.317 & 0.319 \\
 & 50\% & 0.341 & 0.341 & 0.342 & 0.338 & \textbf{0.343} & 0.341 & 0.337 & 0.337 \\
\cline{1-10}
\multirow[t]{3}{*}{80\%} & 10\% & 0.307 & 0.308 & \textbf{0.309} & 0.307 & 0.306 & 0.305 & 0.305 & 0.305 \\
 & 25\% & 0.320 & \textbf{0.321} & 0.321 & 0.320 & 0.318 & 0.319 & 0.317 & 0.319 \\
 & 50\% & 0.342 & 0.341 & 0.342 & 0.340 & \textbf{0.344} & 0.342 & 0.337 & 0.337 \\
\cline{1-10}
\bottomrule
\end{tabular}

    \caption{Policy values calculated from the test set, averaged over 5 iterations varying the missingness seed. Left columns show the percentage of treatments observed and the available treatment budget $\kappa$. Highest policy value in each row is in bold. DR-MAR, DR-CC, and DR-MCCAR all perform similarly, while MTRNET performs well at higher budgets even with treatment observation rates.}
\end{table}

\begin{figure}[H]
    \centering
    \includegraphics[width=0.9\textwidth]{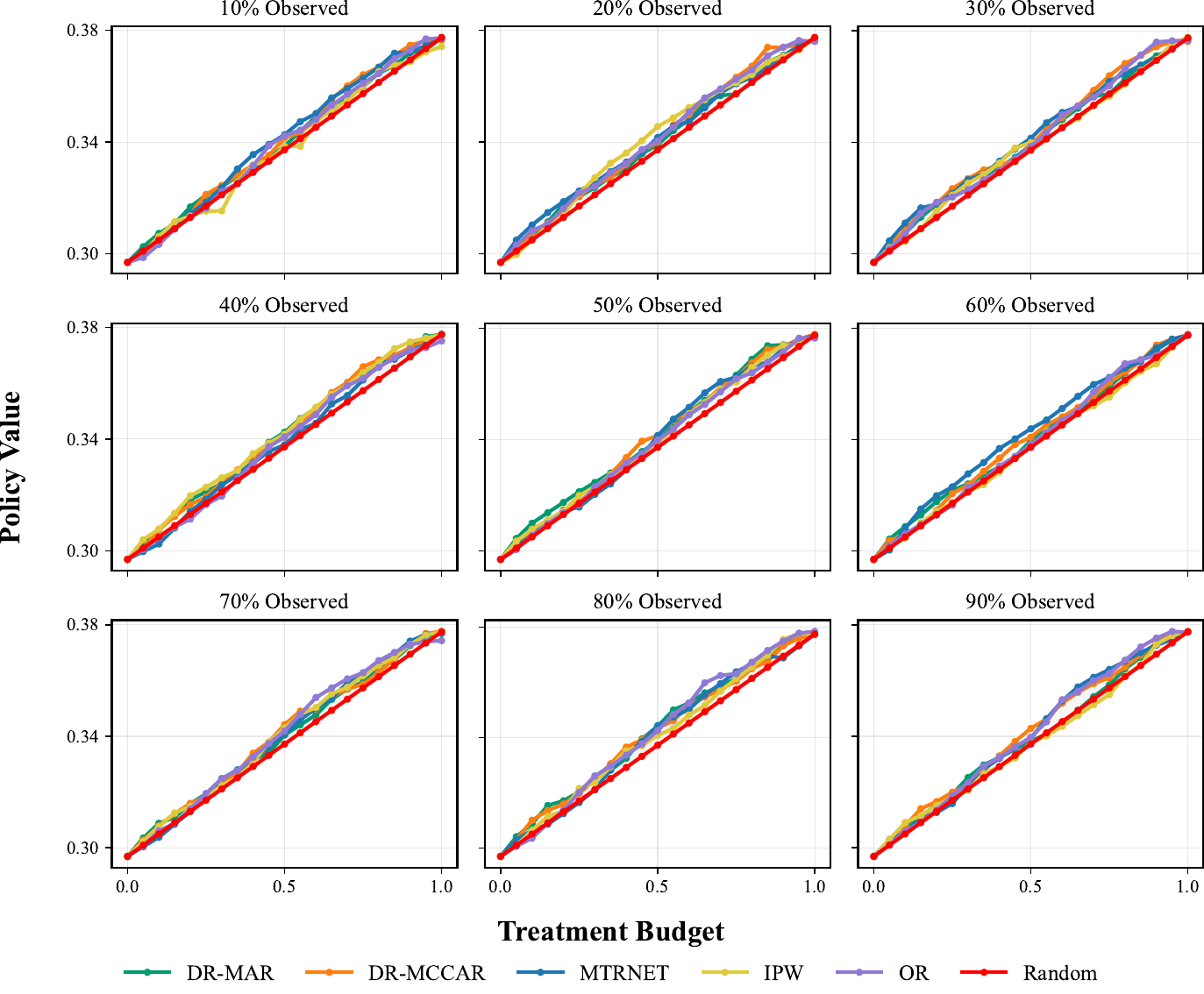}
    \caption{Example policy values over different budgets and treatment observation rates. In the MCCAR setting, all methods are competitive.}
\end{figure}

\subsubsection{STAR - MAR}

\begin{table}[H]
    \centering
    \begin{tabular}{cccccccccc}
\toprule
 &  & DR-CC & DR-MAR & DR-MCCAR & IPW & MTRNET & OR & Random & Risk \\
Obs. Rate & $\kappa$ &  &  &  &  &  &  &  &  \\
\midrule
\multirow[t]{3}{*}{20\%} & 10\% & 1889.8 & \textbf{1899.1} & 1888.6 & 1885.9 & 1886.7 & 1891.4 & 1888.8 & 1890.8 \\
 & 25\% & 1894.4 & \textbf{1905.1} & 1895.5 & 1886.3 & 1889.8 & 1900.2 & 1891.1 & 1895.1 \\
 & 50\% & 1899.3 & \textbf{1909.7} & 1900.2 & 1883.0 & 1897.8 & 1903.2 & 1895.0 & 1898.3 \\
\cline{1-10}
\multirow[t]{3}{*}{50\%} & 10\% & \textbf{1897.6} & 1896.2 & 1893.7 & 1884.3 & 1890.2 & 1894.0 & 1888.8 & 1890.8 \\
 & 25\% & \textbf{1902.4} & 1902.1 & 1902.4 & 1884.2 & 1895.8 & 1901.3 & 1891.1 & 1895.1 \\
 & 50\% & 1906.3 & \textbf{1919.2} & 1907.7 & 1888.8 & 1900.3 & 1908.2 & 1895.0 & 1898.3 \\
\cline{1-10}
\multirow[t]{3}{*}{80\%} & 10\% & 1898.3 & \textbf{1899.1} & 1894.8 & 1888.7 & 1890.8 & 1888.8 & 1888.8 & 1890.8 \\
 & 25\% & 1903.2 & \textbf{1908.3} & 1903.4 & 1890.7 & 1894.3 & 1888.6 & 1891.1 & 1895.1 \\
 & 50\% & 1909.0 & \textbf{1915.9} & 1911.0 & 1895.2 & 1899.9 & 1892.8 & 1895.0 & 1898.3 \\
\cline{1-10}
\bottomrule
\end{tabular}

    \caption{Policy values on the test set, averaged over 10 replications varying the 
    missingness seed. Left columns show the percentage of treatments observed and the available treatment budget $\kappa$. Highest policy value in each row is in bold. DR-MAR outperforms all other methods at low treatment observation rates and achieves the highest overall average policy value.}
\end{table}

\begin{figure}[H]
    \centering
    \includegraphics[width=0.9\textwidth]{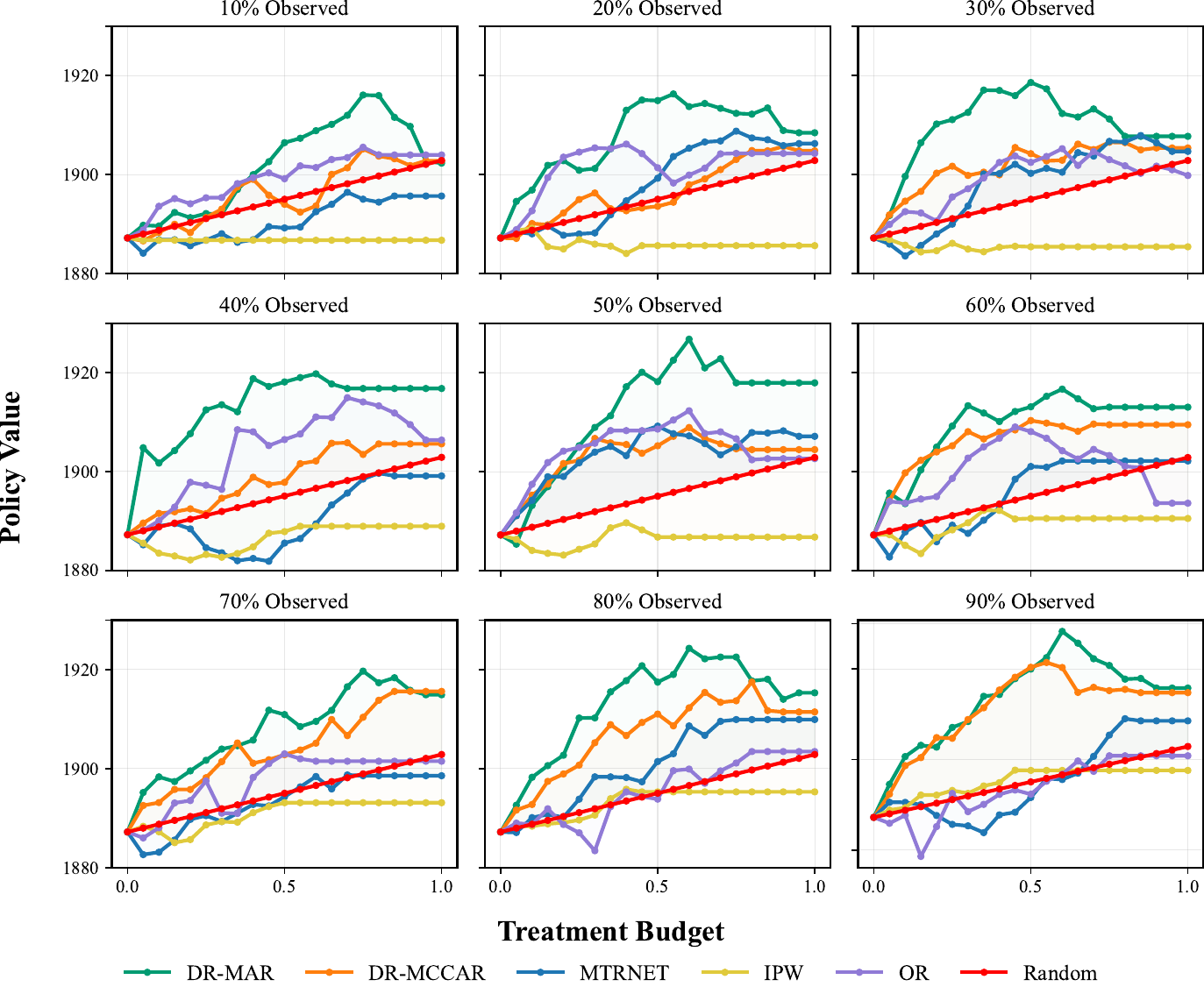}
    \caption{Example policy values across budgets and treatment observation rates.  Even at low treatment observation rates, DR-MAR achieves high policy values. All  DR-methods perform well at high treatment observation rates.}
\end{figure}

\subsubsection{STAR - MCCAR}

\begin{table}[H]
    \centering
    \begin{tabular}{cccccccccc}
\toprule
 &  & DR-CC & DR-MAR & DR-MCCAR & IPW & MTRNET & OR & Random & Risk \\
Obs. Rate & $\kappa$ &  &  &  &  &  &  &  &  \\
\midrule
\multirow[t]{3}{*}{20\%} & 10\% & 1891.5 & \textbf{1892.8} & 1891.2 & 1884.4 & 1887.2 & 1892.4 & 1888.8 & 1890.8 \\
 & 25\% & \textbf{1899.0} & 1896.6 & 1898.6 & 1885.7 & 1890.4 & 1897.2 & 1891.1 & 1895.1 \\
 & 50\% & 1905.6 & 1905.5 & \textbf{1907.2} & 1885.9 & 1891.8 & 1906.7 & 1895.0 & 1898.3 \\
\cline{1-10}
\multirow[t]{3}{*}{50\%} & 10\% & 1895.5 & \textbf{1900.0} & 1898.2 & 1890.0 & 1890.5 & 1895.7 & 1888.8 & 1890.8 \\
 & 25\% & 1904.0 & 1906.0 & \textbf{1907.0} & 1893.4 & 1899.6 & 1904.2 & 1891.1 & 1895.1 \\
 & 50\% & 1916.3 & 1915.2 & 1914.4 & 1896.4 & 1907.6 & \textbf{1917.3} & 1895.0 & 1898.3 \\
\cline{1-10}
\multirow[t]{3}{*}{80\%} & 10\% & 1898.8 & \textbf{1900.3} & 1900.0 & 1890.8 & 1892.0 & 1899.0 & 1888.8 & 1890.8 \\
 & 25\% & 1905.4 & \textbf{1906.4} & 1905.7 & 1893.7 & 1897.6 & 1906.4 & 1891.1 & 1895.1 \\
 & 50\% & 1916.9 & 1917.4 & 1917.0 & 1896.4 & 1906.0 & \textbf{1919.8} & 1895.0 & 1898.3 \\
\cline{1-10}
\bottomrule
\end{tabular}

    \caption{Policy values on the test set, averaged over 10 replications varying the 
    missingness seed. Left columns show the percentage of treatments observed and the available treatment budget $\kappa$. Highest policy value in each row is in bold. Overall, DR-MAR has the most wins, but DR-CC and DR-MCCAR are also competitive in this setting.}
\end{table}

\begin{figure}[H]
    \centering
    \includegraphics[width=0.9\textwidth]{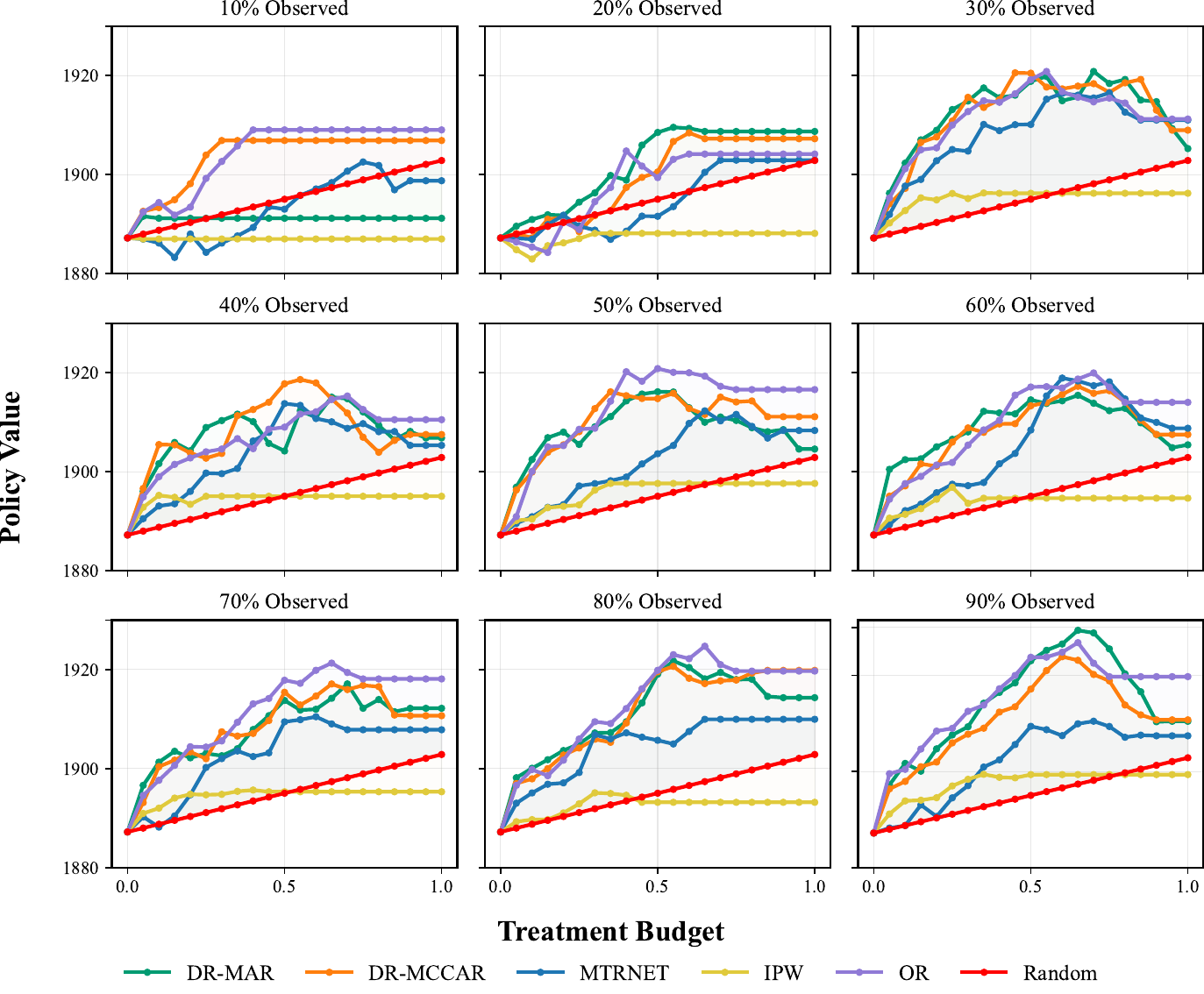}
    \caption{Example policy values over different budgets and treatment observation rates. At treatment observation rates greater than 10\%, all doubly-robust models are competitive.}
\end{figure}

\end{document}